\documentclass{article}

\usepackage[preprint]{neurips_2025}

\usepackage{listings}
\usepackage{siunitx}
\usepackage{enumitem}
\usepackage[utf8]{inputenc} 
\usepackage[T1]{fontenc}    
\usepackage{hyperref}       
\usepackage{url}            
\usepackage{booktabs}       
\usepackage{amsfonts}       
\usepackage{nicefrac}       
\usepackage{microtype}      
\usepackage{xcolor}         
\usepackage{wrapfig} 
\usepackage{tabularx}
\usepackage{booktabs}

\usepackage{amssymb}
\usepackage{pifont}
\newcommand{\cmark}{\ding{51}}  
\newcommand{\xmark}{\ding{55}}  
\usepackage{bm}
\usepackage{multirow}
\usepackage{pifont}
\usepackage{algorithm}
\usepackage{algorithmic}
\usepackage{wrapfig}
\usepackage{colortbl}

\usepackage{amsmath}
\usepackage{amssymb}
\usepackage{mathtools}
\usepackage{amsthm}
\usepackage[most]{tcolorbox}
\usepackage{graphicx}
\usepackage{hyperref,xcolor}

\theoremstyle{plain}

\theoremstyle{definition}

\theoremstyle{remark}

\definecolor{midnightgreen}{rgb}{0.0, 0.29, 0.33}
\definecolor{deepgreen}{HTML}{055c29}
\definecolor{deeppurple}{HTML}{7030a0}
\definecolor{deepblue}{HTML}{171d91}
\definecolor{brown}{HTML}{843c0c}
\definecolor{shadered}{HTML}{ffe5e5}
\definecolor{shadegreen}{HTML}{e5f7ed}
\definecolor{msftBlack}{RGB}{0,0,0}
\definecolor{lightred}{RGB}{255,163,163}
\definecolor{deepred}{RGB}{153,0,0}
\definecolor{aliceblue}{rgb}{0.94, 0.97, 1.0} 
\definecolor{softblue}{RGB}{30, 90, 160}

\hypersetup{
  colorlinks=true,
  linkcolor=deepred,
  citecolor=softblue,
  urlcolor=magenta
}

\newcommand{\methodname}{TimeMaster} 
\title{\methodname: Training Time-Series Multimodal LLMs to Reason via Reinforcement Learning}

\author{
Junru Zhang$^{1}$\thanks{Equal contribution.}\quad
Lang Feng$^{2}$\footnotemark[1]\quad
Xu Guo$^{3}$\quad
Yuhan Wu$^{1}$\quad
Yabo Dong$^{1}$\quad
Duanqing Xu$^{1}$ \\
$^{1}$Zhejiang University, China \\
$^{2}$Nanyang Technological University, Singapore \\
$^{3}$KTH Royal Institute of Technology, Sweden \\
}

\begin{document}

\maketitle

\begin{abstract}

Time-series reasoning remains a significant challenge in multimodal large language models (MLLMs) due to the dynamic temporal patterns, ambiguous semantics, and lack of temporal priors. In this work, we introduce TimeMaster, a reinforcement learning (RL)-based method that enables time-series MLLMs to perform structured, interpretable reasoning directly over visualized time-series inputs and task prompts. TimeMaster adopts a three-part structured output format, \emph{reasoning}, \emph{classification}, and \emph{domain-specific extension}, and is optimized via a composite reward function that aligns format adherence, prediction accuracy, and open-ended insight quality. The model is trained using a two-stage pipeline: we first apply supervised fine-tuning (SFT) to establish a good initialization, followed by Group Relative Policy Optimization (GRPO) at the token level to enable stable and targeted reward-driven improvement in time-series reasoning. We evaluate TimeMaster on the TimerBed benchmark across six real-world classification tasks based on Qwen2.5-VL-3B-Instruct.
TimeMaster achieves state-of-the-art performance, outperforming both classical time-series models and few-shot GPT-4o by over 14.6\% and 7.3\% performance gain, respectively. Notably, TimeMaster goes beyond time-series classification: it also exhibits expert-like reasoning behavior, generates context-aware explanations, and delivers domain-aligned insights. Our results highlight that reward-driven RL can be a scalable and promising path toward integrating temporal understanding into time-series MLLMs. 
\let\thefootnote\relax\footnotetext{Code: \url{https://github.com/langfengQ/TimeMaster}}

\end{abstract}

\section{Introduction}

Time series analysis is a fundamental technique in data mining, enabling the modeling of temporal patterns and informing decision-making across critical domains such as healthcare \cite{zhang2022deep, zhang2024diverse}, industrial monitoring \cite{wang2023sensor, ragab2022conditional}, and environmental surveillance \cite{hu2020understanding}. While the advancements of deep learning have propelled progress in classical time series tasks like forecasting \cite{drouin2022tactis, eldele2024tslanet, zhang2023adacket, wu2022timesnet, husken2003recurrent} and classification \cite{eldele2024tslanet,wu2022timesnet}, the recent surge of large language models (LLMs) \cite{achiam2023gpt, team2023gemini, bai2023qwen, touvron2023llama} invites a broader question: can the rich reasoning abilities of LLMs be extended to the temporal domain, yielding a general-purpose time series analysis that not only predicts but also reasons, explains and advises?

Reasoning is an emergent capability of modern LLMs, enabled by their strong natural language understanding and instruction-following abilities, as represented by Chain-of-Thought Prompting \cite{wei2022chain}.
These strengths provide a natural foundation for moving beyond traditional tasks toward interpretable time-series reasoning (TsR)~\cite{kong2025position}. Despite this, current approaches on time-series multimodal LLMs (MLLMs) face fundamental limitations in achieving deep, robust reasoning. Prompt-based methods~\cite{liu2024picture, jiang2025explainable,merrill2024language}, while flexible and capable of zero- and few-shot generalization, often struggle with the lack of time-series priors and temporal dynamics, even in frontier MLLMs like GPT-4o~\cite{openai2024gpt4o}. This could lead to shallow or inconsistent reasoning over time-series data. On the other hand, post-training strategies, including supervised fine-tuning (SFT) on synthetic~\cite{xie2024chatts, lee2025timecap} or human-annotated~\cite{wang2024news, xu2024beyond, liu2024time, kim2024multi, kong2025time, jia2024gpt4mts, lee2025timecap} time series datasets, can introduce better domain alignment. Yet, these approaches often suffer from limited data diversity, overfitting to template-like patterns, and constrained generalizability. Critically, these datasets with aligned time-series and text pairs often fail to fully capture the breadth of reasoning skills and domain knowledge necessary for robust temporal comprehension. As such, training time-series MLLMs solely on static datasets tends to produce superficial outputs that lack the depth, reliability, and interpretability required for effective time series reasoning.

In this work, we propose \methodname{}, a reinforcement learning (RL)~\cite{sutton2018reinforcement}-based method for multimodal time series reasoning. \methodname{} addresses critical limitations of existing approaches, including dependence on large-scale manually collected datasets, weak reasoning capabilities, and poor performance across diverse time series domains. Instead, \methodname{} learns through iterative RL training, enabling the model to progressively acquire, refine, and generalize its time-series reasoning abilities in a self-improving manner.

\begin{figure}[t]
    \centering
    \includegraphics[width=0.95\textwidth]{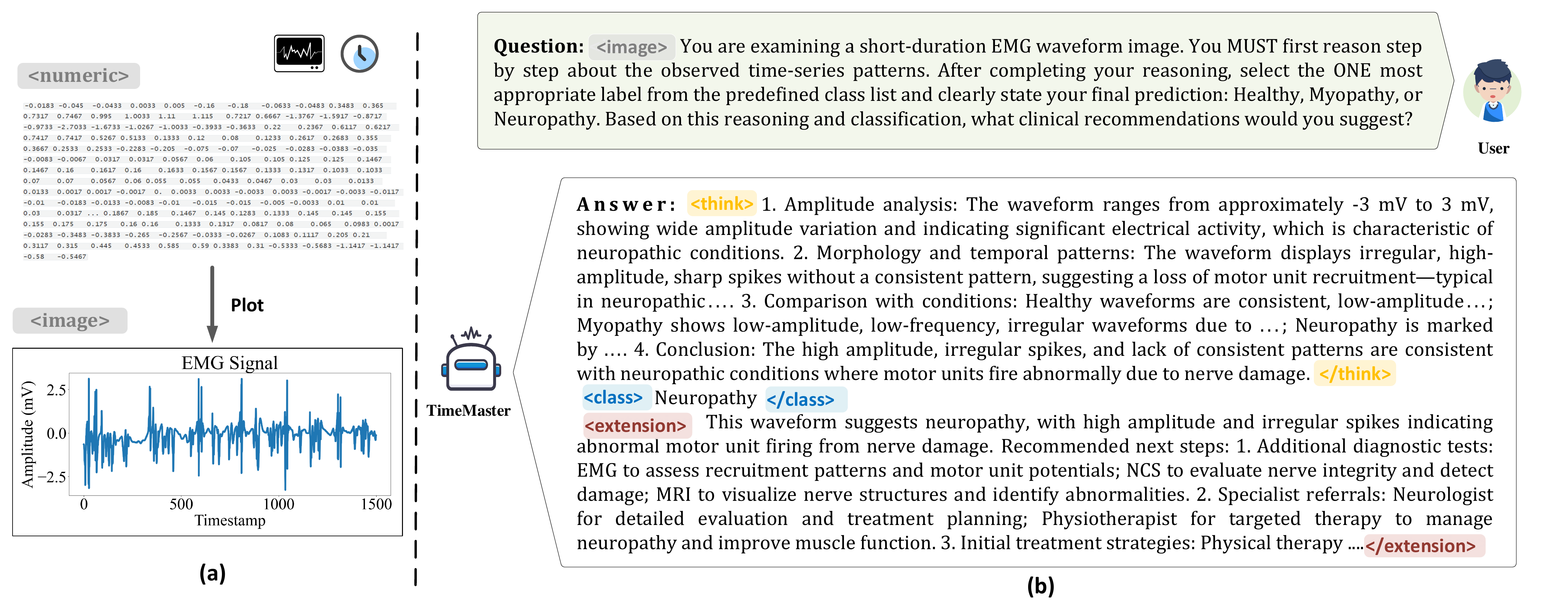} 
    \caption{Example of  \methodname{}. (a) A plot of a raw EMG signal  (1500 time points), labeled 'Neuropathy', from the motor disorder diagnosis domain. (b) The signal image, combined with a task description, is input into \methodname{} for classification, reasoning, and clinical suggestions.}
    \label{fig:example}
\end{figure}
Central to our approach is a structured output format that unifies the model’s response into three components: \textit{reasoning}, \textit{classification}, and \textit{extension}. Given a time series input, the model first generates a coherent chain of thought to explain its analysis, then outputs a classification result, and optionally provides an extension such as suggestions, diagnostics, or actionable insights based on specific task requirements. 
To optimize the model, \methodname{} employs a two-stage learning pipeline. First, we initialize the model with basic task alignment through SFT on a high-quality dataset containing just 1k samples, a fraction of the scale used in prior work (e.g., approximately 200k in \cite{kong2025time}). Then, \methodname{} adopts RL training driven by a composite reward function: a \textit{format reward} to enforce structural consistency, a \textit{hard reward} to ensure classification accuracy, and a \textit{soft reward} to evaluate the quality of open-ended extension. Both the format and hard rewards provide strict binary feedback. The soft reward, due to the open-ended nature of the extension, is computed using an \emph{LLM-as-a-Judge}~\cite{zheng2023judging}. This reward design enables RL to shape both precision and expressiveness in multimodal time series reasoning.  Figure~\ref{fig:example} illustrates an example of \methodname{} applied to a complex  reasoning task on electromyography (EMG) signal classification, highlighting its ability to perform structured reasoning and informed classification.

We evaluate \methodname{} on \textsc{TimerBed} \cite{liu2024picture}, a comprehensive benchmark designed to assess time-series reasoning across six real-world classification tasks.
Empirical results show that \methodname{} achieves state-of-the-art performance, surpassing both traditional time-series baselines and strong multimodal LLMs: it outperforms few-shot GPT-4o by 7.3\% and classical time-series models by over 14.6\% on average. More importantly, \methodname{} exhibits emergent expert-like reasoning behaviors, including iterative self-correction, temporal pattern recognition, and the articulation of causal explanations. These capabilities substantially enhance the interpretability and real-world applicability of MLLMs in temporal domains.
For instance,  in sports monitoring or healthcare scenarios, \methodname{} moves beyond outputting simple labels; it generates rich, human-aligned explanations that reflect domain reasoning, improving both transparency and trustworthiness.
Overall, this work demonstrates a promising step toward equipping multimodal LLMs with time-series reasoning capabilities, facilitating more context-aware, interpretable, and actionable insights in real-world applications.

\section{Related Work}

\paragraph{Time-series foundation models.}
Recent studies have explored diverse paradigms for leveraging large language models (LLMs) in time series analysis, with a particular focus on enhancing predictive performance. A prominent direction \cite{jin2023time, liu2024time, liu2024autotimes, liu2025calf, suntest} involves reprogramming pre-trained LLMs, such as GPT-2 \cite{radford2019language} and LLaMA-7B \cite{touvron2023llama}, as feature extractors by reformulating time series data into textual sequences. This formulation enables direct inference or forecasting by leveraging LLMs' strong sequential modeling abilities.
To support more expressive and context-rich time series analysis, recent work has also emphasized the multimodal nature of real-world applications, integrating intra-series semantics with external contextual signals \cite{liu2025can}. Typical strategies include augmenting time series with domain-relevant textual descriptions \cite{jin2023time, liu2025timecma, liu2024autotimes, liu2025calf}, aligning physiological signals with medical documentation \cite{wan2024meit}, linking stock price trends to financial news \cite{wang2024news, tavakoli2025multi}, and incorporating geographic information \cite{chen2024terra}, traffic records \cite{zhang2024bjtt}, or extra event data \cite{han2024event} into traffic flow modeling.

\paragraph{Reasoning in time-series LLMs.}
With the growing integration of language models into temporal modeling, the field is moving toward a unified reasoning paradigm centered on explanation and causal inference.
Despite its promise, time series reasoning (TsR) remains in an early stage.  Recent efforts \cite{jiang2025explainable, liu2025evaluating, merrill2024language} explore prompting strategies tailored to structured reasoning in time series tasks. For example, VL-Time \cite{liu2024picture} introduces visualized time series directly into GPT-4o for zero-shot or few-shot reasoning, offering an alternative to purely numerical inputs.
Nevertheless, such methods are constrained by the absence of temporal priors and domain-specific inductive biases in standard LLM pretraining. To address these limitations, some approaches adopt post-training strategies using synthetic time-series–text paired corpora. For example, Time-MQA \cite{kong2025time} aggregates 200k time-series–text pairs annotated by GPT-4o across various domains for supervised fine-tuning. However, constructing such datasets remains resource-intensive, often relying on manual schema design or commercial LLM APIs. This poses  challenges to encompassing broad domain knowledge and capturing the full diversity of real-world temporal phenomena.
Our work proposes a cost-efficient yet effective strategy to equip multimodal LLMs with time-series reasoning capabilities, using only a small amount of task-specific supervision.

\paragraph{RL in large language models.}
Reinforcement learning (RL)~\cite{sutton2018reinforcement} has emerged as a powerful framework for advancing the capabilities of LLMs. One of the most influential approaches in this domain is Reinforcement Learning from Human Feedback (RLHF)~\cite{ziegler2019fine,stiennon2020learning,ouyang2022training,rafailov2024direct}, which introduces human preference as a guiding signal during the fine-tuning stage. By optimizing model outputs against reward models trained on human judgments, RLHF enables LLMs to generate responses that are more aligned with human expectations in terms of helpfulness, coherence, and ethical appropriateness. 
In addition, RL is increasingly used to enhance the reasoning abilities of LLMs~\cite{team2025kimi,shao2024deepseekmath,yu2025dapo}. Recent work, such as DeepSeek-R1~\cite{guo2025deepseek}, incorporates RL objectives to encourage logical reasoning and structured problem-solving. Instead of solely optimizing for linguistic fluency, these models are trained to maximize rule-based rewards, such as correctness in mathematical derivations, enabling LLMs to perform complex reasoning tasks with greater accuracy and internal consistency. Given the growing demand for reliable decision-making in the time-series domain, we propose leveraging the potential of RL by developing strategies that address key gaps in time-series reasoning with multimodal LLMs.

\section{\methodname{}}
Time-series data, collected from diverse sources such as scientific experiments, industrial systems, and healthcare monitoring, underpins a wide range of classification tasks critical to real-world decision-making. 
Beyond conventional classification, practical decision-making often demands models capable of \textit{reasoning} over temporal patterns, articulating causal relationships, and proposing actionable insights. While recent time-series models~\cite{wu2022timesnet, liu2023itransformer,liu2023itransformer,nie2022time} have achieved substantial progress, they predominantly focus on optimizing numeric prediction objectives, often at the expense of interpretability and explainability. Meanwhile, LLMs such as ~\cite{touvron2023llama, bai2023qwen} excel in natural language generation but are not inherently equipped for structured temporal reasoning, limiting their effectiveness in understanding and reasoning about the complex dynamics of real-world time series.

In this section, we present \methodname{}, an RL-enhanced time-series MLLM that acquires expert-level reasoning over time-series signals. It accepts multimodal inputs and outputs a structured explanation, classification, and optional extension, optimized for both accuracy and interpretability.

\subsection{Multimodal Input}
Time series reasoning refers to the open-ended and structured capability of an MLLM to interpret and analyze temporal patterns by jointly considering raw time series data $\mathbf{X}$ and auxiliary contextual information $\mathbf{q}$ for classification tasks. Formally, let $\mathbf{X} = \{ \mathbf{x}_t \}_{t=1}^T$ denote a multivariate time series of length $T$, where each $\mathbf{x}_t \in \mathbb{R}^D$ represents a $D$-dimensional observation at time step $t$. Let $\mathbf{q} = \{ q_1, q_2, \dots\}$ be a sequence of textual tokens, such as natural language instructions or domain knowledge, where each $q_i \in \mathcal{V}_{\text{text}}$ is sampled from a textual vocabulary specific to the MLLM. 
Unlike traditional approaches~\cite{jin2023time,liu2024time, liu2024autotimes, liu2025calf, suntest, wu2022timesnet, liu2023itransformer} that focus solely on predictions, our formulation of time series reasoning empowers the model to uncover latent temporal structures, infer causal relationships, and generate interpretable insights \emph{through natural language}.  It bridges low-level numerical signals with high-level reasoning objectives within a unified, multimodal framework.

We formally define a time-series MLLM as $\pi_\theta$ parameterized by $\theta$, which specifies a conditional probability distribution over the output sequence $\mathbf{y} = \{ y_1, y_2, \dots, y_K \}$, given the inputs $\mathbf{X}$ and $\mathbf{q}$:
\begin{equation}
\pi_\theta(\mathbf{y} \mid \mathbf{X}, \mathbf{q}) = \prod\nolimits_{k=1}^K \pi_\theta(y_k \mid y_{<k}, \mathbf{X}, \mathbf{q}).
\end{equation}
The autoregressive factorization enables the MLLM to generate structured, context-aware outputs by attending to multimodal inputs and prior outputs sequentially.
To reduce token cost and enhance perceptual efficiency, we adopt the visualization-based input strategy proposed by VL-Time~\cite{liu2024picture}, wherein the time series $\mathbf{X}$ is transformed into an image representation (a line plot), as illustrated in Fig.~\ref{fig:example} (a). These visualized inputs are then processed by the MLLM, leveraging its visual reasoning capabilities while maintaining an intuitive, compact, and semantically rich input format \cite{xu2025beyond}.

\subsection{Structured Output for Time-Series Reasoning}
Rather than producing a single label or forecast, \methodname{} generates a \textit{textual} output formatted in a structured manner using special XML-style tags. The full output is a sequence of natural language tokens organized into three defined segments:
\begin{equation}
\underbrace{\langle \texttt{think}\rangle\;\cdots\;\langle/\!\texttt{think}\rangle}_{\text{Reasoning}}
\quad
\underbrace{\langle \texttt{class}\rangle\;\cdots\;\langle/\!\texttt{class}\rangle}_{\text{Classification}}
\quad
\underbrace{\langle \texttt{extension}\rangle\;\cdots\;\langle/\!\texttt{extension}\rangle}_{\text{Optional Extension}}.
\end{equation}
The \texttt{think} block expresses open-ended time-series reasoning in a ``chain-of-thought''~\cite{wei2022chain} style. This includes pattern recognition, trend analysis, causal inference, or domain-specific signal interpretation. For example, in an EMG signal, the model might describe amplitude variation, waveform shape, and frequency shifts before concluding.
The \texttt{class} block outputs a discrete label, indicating the predicted class (e.g., \texttt{Normal}, \texttt{Myopathy}, \texttt{Neuropathy}). This enables rigorous evaluation against ground truth during training.
The \texttt{extension} block optionally provides follow-up insights or suggestions, such as diagnostics, forecasts, remediation plans, or next-step actions. This facilitates practical usability, especially in domains like medicine and industrial monitoring.

\begin{figure*}[t]
    \centering
    \includegraphics[width=1\textwidth]{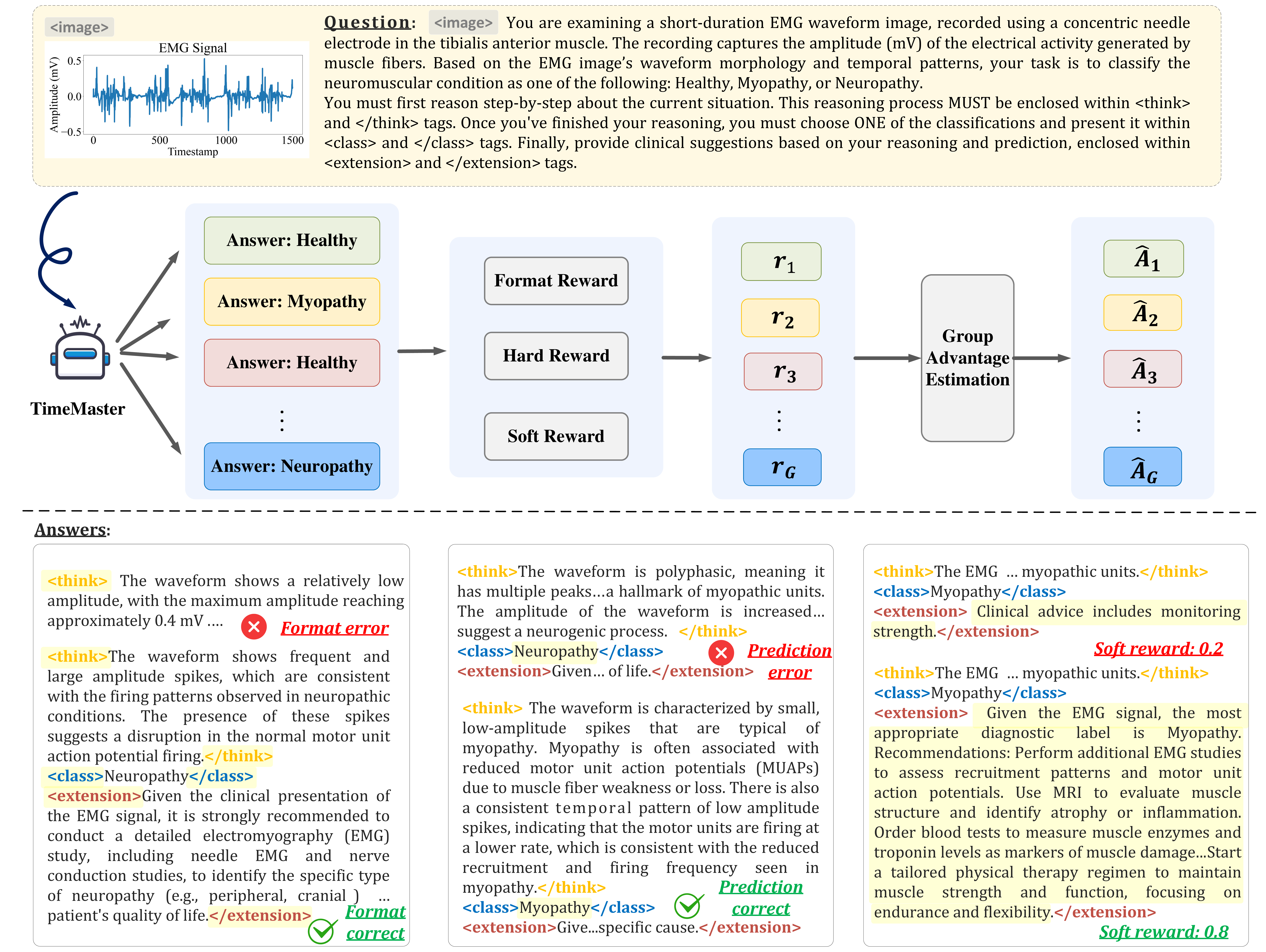} 
    \caption{ Overview of the proposed \methodname{}. }
    \label{fig:overview}
\end{figure*}

\subsection{Reward Modeling}
Training \methodname{} to generate high-quality reasoning outputs requires balancing multiple objectives simultaneously: structural correctness, classification accuracy, and the open-ended relevance of advice. To impose these multifaceted requirements, we construct a \emph{composite reward} that decomposes along three axes: \emph{format reward}, \emph{hard reward}, and \emph{soft reward}. Each targets a distinct dimension of output quality, as visualized in Figure~\ref{fig:overview}.

\textbf{Format reward $r^{\mathrm{fmt}}$.}~~
The format reward enforces strict adherence to the predefined output structure. A regex-based parser automatically checks if the generated response contains correctly nested and ordered \texttt{<think>}, \texttt{<class>}, and \texttt{<extension>} tags, with non-empty reasoning and classification spans. If the format is fully correct, a binary reward of $r^{\mathrm{fmt}}(\mathbf{y}) = 1$ is assigned; otherwise, $r^{\mathrm{fmt}}(\mathbf{y}) = 0$. Figure~\ref{fig:overview} (left) demonstrates a case of format reward, where clinically valid reasoning is penalized if the tag structure is broken (format error), ensuring that the model maintains syntactic discipline throughout training.

\textbf{Hard reward $r^{\mathrm{hard}}$.}~~
The hard reward evaluates the factual correctness of the model's classification results. Given ground-truth labels $c^\star\in\mathcal{C}$, an exact match between the predicted \texttt{<class>} tag and the true label grants a reward of $r^{\mathrm{hard}} = 1$; any mismatch yields a reward of $r^{\mathrm{hard}} = 0$. Formally, let $\hat{c}(\mathbf{y})\in\mathcal{C}$ denote the class extracted from the \texttt{<class>} tag, we define
\begin{equation}
\label{eq:reward_hard}
r^{\mathrm{hard}}\bigl(\hat{c},c^\star\bigr) = \mathbb{I}[\hat{c}=c^\star],
\end{equation}
where $\mathbb{I}[\cdot]$ is the indicator function. This hard reward ensures that correct reasoning steps are aligned with accurate final classification. Figure~\ref{fig:overview} (middle) illustrates this design: although the reasoning mentions myopathy features, the model incorrectly predicts \texttt{Neuropathic}, leading to a prediction error and a zero hard reward.

\textbf{Soft reward $r^{\mathrm{soft}}$.}~~
As the content of \texttt{<extension>}, such as suggestions, diagnostics, or actionable insights tailored to specific task requirements, is inherently open-ended, there is often no single definitive notion of right or wrong. To robustly assess such outputs, we introduce a soft reward grounded in the \textit{LLM-as-a-Judge} paradigm~\cite{zheng2023judging}.
Specifically, we utilize GPT-4o~\cite{openai2024gpt4o}, prompted to evaluate each extension along four dimensions: (\textbf{1}) \textit{Specificity}: Assess whether the extension is tailored to the given context, rather than offering generic or broadly applicable suggestions. (\textbf{2}) \textit{Appropriateness}: Determine whether the extension is reasonable and well-aligned with the preceding reasoning and prediction. (\textbf{3}) \textit{Relevance}: Evaluate whether the extension logically follows from the prior context, avoiding fallback or disconnected recommendations. (\textbf{4}) \textit{Depth}: Judge whether the extension reflects domain-specific understanding, such as suggesting detailed next steps, considerations, or nuanced insights. Formally, let $\phi$  denote the LLM judge and $e(\mathbf{y})$ denote the extension content extracted from the \texttt{<extension>} tag, we define the soft reward as
\begin{equation}
\label{eq:reward_soft}
r^{\mathrm{soft}}(e(\mathbf{y}),c^\star) = \mathbb{I}[\hat{c}=c^\star]\cdot\mathrm{mean}\{s^{\text{spec}},s^{\text{appr}},s^{\text{rel}},s^{\text{dep}}\},  \ \text{where} \ \{s^{\text{spec}},s^{\text{appr}},s^{\text{rel}},s^{\text{dep}}\} = \phi(e(\mathbf{y}), c^\star).
\end{equation}
Each dimension is rated on a continuous scale $[0, 1]$, and the mean score is used as the soft reward, conditional on correct classification.
As illustrated in Figure~\ref{fig:overview} (right), even when the primary classification is correct, an extension that is generic or vague (e.g., ``monitor strength'') receives a low soft reward. In contrast, extensions that are specific and actionable (e.g., ``recommend blood tests to measure muscle enzymes and troponin levels'') achieve higher scores, better reflecting their clinical appropriateness and practical utility.

\textbf{Composite reward.}~~
Finally, we combine the three reward components into an overall reward signal to guarantee format fidelity, classification correctness, and extension utility of time-series reasoning:
\begin{equation}
\label{eq:total_reward}
r = \lambda^{\mathrm{fmt}}\,r^{\mathrm{fmt}} + \lambda^{\mathrm{hard}}\,r^{\mathrm{hard}} + \lambda^{\mathrm{soft}}\,r^{\mathrm{soft}}.
\end{equation}
where $\lambda^{\mathrm{fmt}}, \lambda^{\mathrm{hard}}, \lambda^{\mathrm{soft}} \geq 0$, trading off the relative importance of each term. This composite reward enables \methodname{} to optimize jointly for output format precision, decision-making accuracy, and time-series reasoning expressiveness, which is essential for robust, interpretable time-series analysis in practical domains.

\subsection{Optimization for Time-Series Reasoning}
To enable \methodname{} to generate high-quality, structured reasoning over time-series inputs, we adopt a two-stage training pipeline. We begin with SFT on a low-resource dataset consisting of 1k high-quality examples sampled from GPT-4o outputs \footnote{See Appendix~\ref{warm-up} for the collected details and cases used in training.}. This stage provides a strong initialization by aligning the model with the desired structured output format and domain semantics.
We then proceed to RL, where we optimize the model using token-level Group Relative Policy Optimization (GRPO)~\cite{shao2024deepseekmath,guo2025deepseek}, which leverages group-wise statistics to normalize advantage estimates without requiring a separate value network.

Formally, for each training context $(\mathbf{X},\mathbf{q})$, we sample a group of $G$ complete output sequences $\{\mathbf{y}_i\}_{i=1}^G \sim \pi_{\theta_{\text{old}}}(\cdot \mid \mathbf{X}, \mathbf{q})$, and compute their scalar rewards $\{r_i\}_{i=1}^G$ via the composite reward function defined in Equation~\eqref{eq:total_reward}. We then calculate the group mean $\mu_r$ and standard deviation $\sigma_r$, and normalize the advantages as:
\begin{equation}
\label{eq:mean_std}
\hat A_i =\frac{r_i-\mu_r}{\sigma_r},\quad \mu_r=\frac{1}{G}\sum_{j=1}^G r_j,\quad \sigma_r=\sqrt{\frac{1}{G}\sum_{j=1}^G(r_j-\mu_r)^2+\varepsilon},
\end{equation}
Each token $y_{i,k}$ in sequence $\mathbf{y}_i$ shares the same normalized advantage $\hat A_i$.
This normalization ensures that advantage estimates remain on a comparable scale across different contexts.

We then update the parameters $\theta$ of time-series MLLM by maximizing the following clipped surrogate objective with an explicit KL-divergence regularization against a frozen reference MLLM $\pi_{\mathrm{ref}}$:
\begin{equation} \label{eq:grpo_token} 
\mathcal{L}(\theta) = \frac{1}{G}\sum_{i=1}^G \frac{1}{|\mathbf{y}_i|} \sum_{k=1}^{|\mathbf{y}_i|}
\min\bigl( \rho_{i,k} \hat A_{i},\; \mathrm{clip}(\rho_{i,k}, 1-\epsilon, 1+\epsilon) \hat A_{i} \bigr)
- \beta\, \mathrm{KL}\bigl[\pi_\theta \parallel \pi_{\mathrm{ref}}\bigr],
\end{equation}
where $\rho_{i,k} = \frac{\pi_\theta(y_{i,k} \mid y_{i,<k}, \mathbf{X}, \mathbf{q})}{\pi_{\mathrm{old}}(y_{i,k} \mid y_{i,<k}, \mathbf{X}, \mathbf{q})}$ is the importance sampling ratio for each generated token, $\epsilon$ is the clipping threshold, and $\beta$ controls the strength of KL regularization. We provide the pseudo code of the training \methodname{} in Appendix~\ref{app:pseudo}.

\begin{table}[t]
  \centering
  \caption{Accuracy (\%) of different methods on \textsc{TimerBed}. ``Simple Determ.'' denotes simple deterministic reasoning and ``Complex Determ.'' denotes complex deterministic reasoning.}
  \label{tab:main}
  \resizebox{\textwidth}{!}{
  \begin{tabular}{lll c c c c c c c}
    \toprule
    \multirow{2}{*}{\textbf{Modality}} & \multirow{2}{*}{\textbf{Type}} & \multirow{2}{*}{\textbf{Method}} &
    \multicolumn{2}{c}{\textbf{Simple Determ.}} & 
    \multicolumn{2}{c}{\textbf{Complex Determ.}} & 
    \multicolumn{2}{c}{\textbf{Probabilistic}} & 
    \multirow{2}{*}{\textbf{Avg.}} \\
    \cmidrule(r){4-5} \cmidrule(r){6-7} \cmidrule(r){8-9}
     & & & \textbf{RCW} & \textbf{TEE} & \textbf{ECG} & \textbf{EMG} & \textbf{CTU} & \textbf{HAR} & \\
    \midrule

    \multirow{8}{*}{Numeric} 
    & \multirow{8}{*}{Classical} 
    & Transformer     & 64.12 & 59.52 & 25.00 & 86.67 & 59.20 & 87.26 & 63.63 \\
    & & Autoformer      & 62.59 & 26.19 & 23.95 & 46.67 & 67.20 & 75.04 & 50.27 \\
    & & Informer        & 75.51 & 59.52 & 22.39 & 66.66 & 67.20 & 85.83 & 62.85 \\
    & & FEDformer       & 76.59 & 42.86 & 26.40 & 73.33 & 51.60 & \textbf{89.88} & 60.11 \\
    & & PatchTST        & 82.11 & 57.14 & 24.82 & 60.00 & 64.00 & 79.60 & 61.28 \\
    & & iTransformer    & 76.92 & 21.43 & 24.48 & 46.67 & 46.40 & 89.49 & 50.90 \\
    & & TimesNet        & 80.23 & 61.90 & 26.20 & 73.33 & 64.00 & 88.65 & 65.72 \\
    & & DLinear         & 56.96 & 47.63 & 23.61 & 46.67 & 52.40 & 48.97 & 46.04 \\
    \midrule

\multicolumn{10}{l}{\emph{Base: GPT-4o}} \\
Numeric+Text & Prompting & GPT-4o (Zero-shot) & 50.00 & 21.43 & 25.00 & 33.33 & 45.45 & 29.17 & 34.06 \\
Image+Text   & Prompting & VL-Time (Zero-shot)   & 70.02 & 24.88 & 26.33 & 33.33 & 50.71 & 37.50 & 40.46 \\
Numeric+Text & Prompting & GPT-4o (Few-shot) & 50.00 & 35.71 & 31.25 & 33.33 & 50.00 & 12.50 & 35.47 \\
Image+Text   & Prompting & VL-Time (Few-shot) & \textbf{91.03} & 64.29 & {43.75} & {91.67} & 63.64 &  {66.67} & {70.18} \\

    \midrule

    \multicolumn{10}{l}{\emph{Base: Qwen2.5-7B-Instruct}} \\
    Numeric+Text & Training & Time-MQA & 36.84 & 10.48 & 25.00 & 18.94 & 38.40 & 16.83 & 24.42 \\
    \midrule

    \multicolumn{10}{l}{\emph{Base: Qwen2.5-VL-3B-Instruct}} \\
    Image+Text & Prompting & Qwen2.5-VL & 47.66 & 13.70 & 20.00 & 17.03 & 46.40 & 16.49 & 26.88 \\
    Image+Text & Training  & Qwen2.5-VL (SFT) & 49.29 & 19.18 & 21.92 & 34.15 & 50.00 & 21.95 & 32.75 \\
    \rowcolor{gray!15} Image+Text & Training  & \textbf{TimeMaster (RL)}  & 72.53& 13.70 & 25.00 & 48.78 & 54.00 & 34.55 & 41.43 \\
    \rowcolor{gray!15} Image+Text & {Training} & \textbf{TimeMaster (SFT+RL)} & {75.56} & \textbf{68.49} & \textbf{60.00} & \textbf{100.00} & \textbf{84.40} & 63.29 & \textbf{75.29} \\
    \bottomrule
  \end{tabular}
  }
\end{table}

\section{Experiment} \label{sec:exp}

\paragraph{Datasets}\label{subsec:timerbed}
We evaluate our model on \textsc{TimerBed}~\citep{liu2024picture}, a recent benchmark for \emph{time-series reasoning} (TsR), comprising six real-world classification datasets grouped by three reasoning patterns: \textbf{1)} Simple deterministic reasoning: RCW detects whale calls based on distinct "whoop" sounds in audio; TEE classifies satellite-captured electromagnetic patterns for space event recognition; \textbf{2)} Complex deterministic reasoning: ECG diagnoses arrhythmias via waveform analysis; EMG detects muscle disorders from neuromuscular signals and \textbf{3)} Probabilistic reasoning: HAR predicts physical activities (e.g., walking) from sensor data; CTU infers device type from electricity usage patterns.
Unlike conventional benchmarks focused solely on accuracy, \textsc{TimerBed} requires models to \emph{reason} and \emph{explain} their predictions, emphasizing interpretability across domains like biology and healthcare.  Detailed descriptions and corresponding prompts are provided in Appendices \ref{app:data} and \ref{app:prompt}.

\paragraph{Baselines}
We compare \methodname{} against a wide range of baselines, following the evaluation setup in \textsc{TimerBed}~\cite{liu2024picture}:
\textbf{1)} Classical Time-Series Models: We evaluate eight strong baselines with diverse architectures (e.g., transformer-based models), including: {Transformer}~\cite{vaswani2017attention}, {Autoformer}~\cite{wu2021autoformer}, {Informer}~\cite{zhou2021informer}, {FEDformer}~\cite{zhou2022fedformer}, {PatchTST}~\cite{nie2022time}, {iTransformer}~\cite{liu2023itransformer}, {TimesNet}~\cite{wu2022timesnet}, and {DLinear}~\cite{zeng2023transformers}.
\textbf{2)} LLM-based Reasoning Methods: Following a recent survey on time-series reasoning with LLMs~\cite{liu2025can}, we evaluate several representative approaches. These include {GPT-4o}, which processes raw numerical time-series inputs~\cite{liu2024picture}, and {VL-Time}~\cite{liu2024picture}, which leverages visualized time-series data. Both are evaluated under zero-shot and few-shot prompting settings. We also include {Time-MQA}~\cite{kong2025time}, a recent TsR model fine-tuned on 200k time-series–text pairs with Qwen2.5-7B-Instruct.

\paragraph{Training Setups}\label{subsec:impl}
We use Qwen2.5-VL-3B-Instruct~\cite{bai2025qwen2} as the base time-series MLLM. Time-series inputs are visualized as RGB plots following the {VL-Time}~\cite{liu2024picture} recipe and paired with task-specific textual prompts.
\textbf{1)} In the SFT phase, we use GPT-4o (temperature = 1.0) to generate around 1k samples per dataset for each task. These samples are used to adapt the model to time-series reasoning via instruction tuning.
\textbf{2)} In the RL phase, we optimize the model using GRPO with group size $G = 5$, a clipping threshold $\epsilon = 0.2$, and KL penalty weight $\beta = 0.001$. The reward is computed using Eq.~\ref{eq:total_reward} with weights $\lambda^{\mathrm{fmt}}=0.1$ and $\lambda^{\mathrm{hard}}=0.9$, and we set $\lambda^{\mathrm{soft}} = 1.0$ in case studies to encourage extension contents.
We evaluate the base Qwen2.5-VL, its SFT version, i.e., Qwen2.5-VL (SFT), our RL-trained version, i.e., TimeMaster (RL), and our full model, i.e., TimeMaster(SFT+RL), which combines both SFT and RL. 
More details are provided in Appendix~\ref{app:imp}.


\subsection{Main Results}\label{subsec:main_results}
Table~\ref{tab:main} presents the performance of different methods on the \textsc{TimerBed} benchmark \cite{liu2024picture}. Our proposed method, \methodname{}, achieves a new state-of-the-art in time-series reasoning with multimodal LLMs, attaining an average accuracy of 75.29\%, which surpasses classical time-series models and few-shot GPT-4o baselines by 14.6\% and 7.3\%, respectively. Notably, this performance is achieved with only a 3B-parameter model, demonstrating significantly greater parameter efficiency compared to GPT-4o  and Qwen2.5-7B-Instruct.
We further observe that TimeMaster (RL), even without SFT, achieves performance comparable to GPT-4o with {VL-Time} zero-shot prompting. This highlights the effectiveness of RL in improving reasoning quality without relying on massive pretraining. In contrast, Time-MQA underperforms across all tasks, likely due to a distribution gap between its training data and the test data in \textsc{TimerBed}, as well as the difficulty in handling long numerical time-series inputs.
However, on HAR, our model slightly lags behind classical methods, likely due to the complex inter-variable dependencies in sensor data. This suggests that incorporating fine-grained temporal reasoning supervision could be a promising direction for future improvements.
Overall, the results confirm that \methodname{} delivers strong classification performance with high accuracy and exceptional parameter efficiency.

\subsection{Key Insights from Empirical Evaluations}\label{sec:insights}
In this section, we distill key insights from our some empirical findings to highlight the core challenges of TsR with MLLMs and to demonstrate the effectiveness of \methodname{} in addressing them.

\begin{figure*}[t]
    \centering
    \includegraphics[width=1\textwidth]{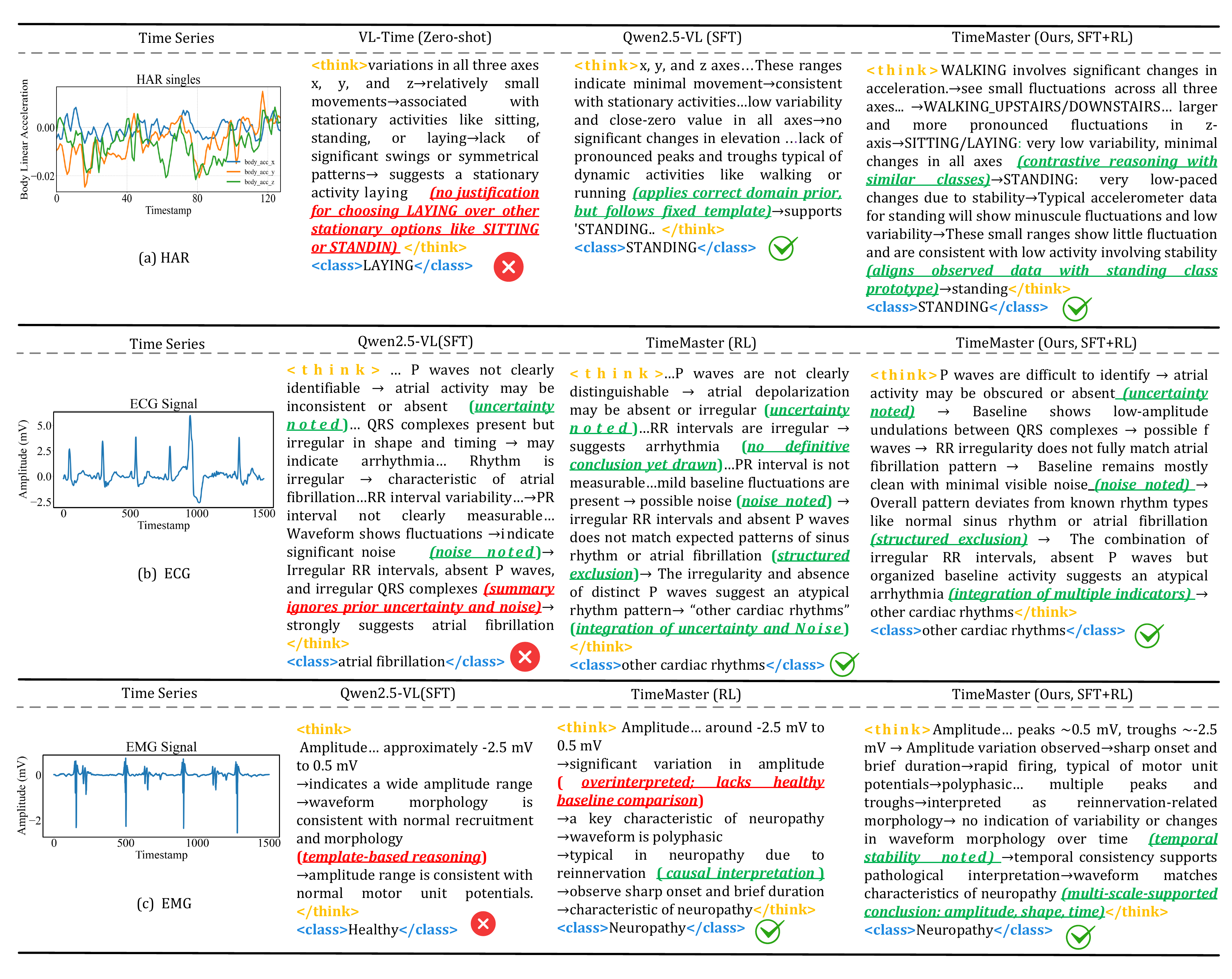} 

    \caption{Comparison of various models' reasoning on three time-series tasks. Shown are key reasoning outputs (Appendix~\ref{app:detail} provides the complete model outputs); italic text in parentheses provides commentary and is not generated by the models.}
    \label{fig:reason}
    \vspace{-1em}
 
\end{figure*}

\paragraph{Takeaway~I: \emph{Training is Essential for Structured Time-Series Reasoning.}}\label{insight1}
We begin by comparing the reasoning chains of three models, i.e., VL-Time (Zero-shot), Qwen2.5-VL (SFT), and TimeMaster (SFT+RL), as shown in Fig.~\ref{fig:reason}(a). The results show that trained Qwen2.5-VL (SFT) and TimeMaster (SFT+RL) exhibit clear, structured reasoning patterns that map latent temporal cues to domain concepts. For example, in the HAR case labeled standing (Fig.~3a), Qwen2.5-VL (SFT) detects low variance across accelerometer axes, links it to a static-posture pattern, and selects standing accordingly. The full TimeMaster model goes further: it performs contrastive elimination of dynamic activities, explicitly mentions the symmetry of micro-fluctuations, and justifies its decision with a causal chain that mirrors expert diagnostic logic. This demonstrates the RL stage’s ability to refine decision-making beyond the SFT prior.
But {VL-Time}, despite using GPT-4o, offers only a shallow description and prematurely predicts laying. These findings underscore that training is essential for incorporating task-specific time-series priors and enabling structured, context-aware TsR.

\paragraph{Takeaway~II: \emph{RL Unlocks Expert-Level Time-Series Reasoning.}}\label{insight2}
We evaluate three models on two physiological signals: an ECG labeled as {other cardiac rhythms} (Fig.~\ref{fig:reason}(b)) and an EMG labeled as {neuropathy} (Fig.~\ref{fig:reason}(c)). Qwen2.5-VL (SFT) shows basic feature recognition, such as identifying absent P waves or amplitude variations. However, its reasoning remains shallow and template-driven, leading to misclassifications. For example, it predicts the ECG as atrial fibrillation based on “irregular RR intervals”  while ignoring noted noise and uncertainty, and mislabels the EMG as healthy using the amplitude range alone.
TimeMaster (RL) demonstrates improved diagnostic reasoning by incorporating contextual cues and prototype deviation analysis. In the ECG case, it integrates multiple uncertain cues (e.g., possible noise, irregular RR intervals) to rule out standard arrhythmia classes. For the EMG, it captures both morphological and temporal markers, such as  sharp onset and brief duration, to support a neuropathic interpretation.
Yet, pure RL may be limited by slow exploration and convergence due to the lack of temporal priors. TimeMaster (RL+SFT) mitigates this by using lightweight SFT as a warm-up to inject basic domain knowledge. This provides a good initialization for the follow-up RL training, steers RL toward clinically salient dimensions early, and improves convergence and stability. As shown, RL with warm-up further facilitates the emergence of structured diagnostic reasoning via reward-guided exploration, alleviating dependence on predefined templates and enabling expert-level interpretations through multi-scale pattern synthesis (e.g., amplitude, waveform morphology, and temporal stability) into a cohesive reasoning framework.

\begin{wrapfigure}{r}{0.49\textwidth}
    \centering
    \vspace{-1em} 
    \includegraphics[width=0.48\textwidth]{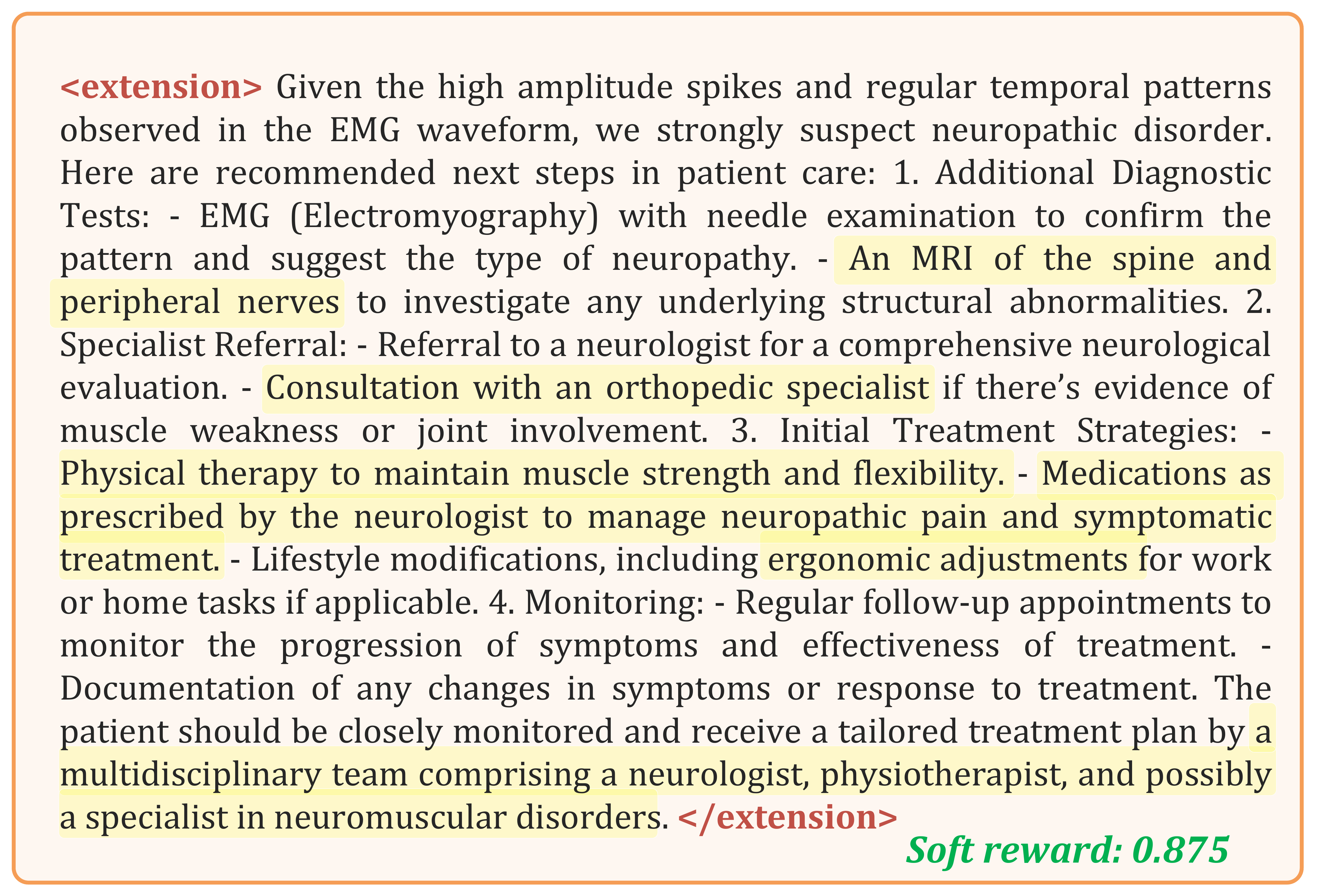}
    \caption{Extension output of \methodname{} via \texttt{<extension>} tags on a neuropathic EMG signal.}
    \label{fig:ext}
   
\end{wrapfigure}

\paragraph{Takeaway III: \emph{Beyond Reasoning, RL Enables Flexible Extensions such as Insightful, Actionable Suggestion Generation.}}

As shown in Figure~\ref{fig:ext}, \methodname{} not only excels at core tasks such as reasoning and classification, but also acquires flexible, value-added capabilities, such as generating high-quality, domain-appropriate clinical suggestions, driven by soft rewards.
In the neuropathy case, its recommendations are specific (e.g., spinal MRI), appropriate (well-aligned with the diagnosis), relevant (logically grounded in the preceding reasoning), and deep (covering diagnostics, specialist referral, treatment planning, and longitudinal monitoring). This reflects an emergent capacity for structured, context-aware reasoning aligned with real-world clinical workflows.
This capability arises from our reward-shaped RL. The soft reward, evaluated via an LLM-as-a-Judge, explicitly encourages suggestions that are coherent, relevant, domain-appropriate, and practically actionable. Through RL, \methodname{} internalizes these preferences and progressively acquires the ability to perform temporal reasoning for downstream clinical tasks. This contrasts with models that rely on fixed heuristics or shallow pattern recognition.
Overall, these results demonstrate that our design enables multimodal LLMs to move beyond classification and basic reasoning toward comprehensive, flexible, expert-level generation of time-series insights in complex decision-making scenarios.

\section{Conclusion}

In this work, we present TimeMaster, a novel RL-enhanced time-series MLLMs designed for structured reasoning over visualized  time-series data. Our design integrates a two-stage training pipeline: beginning with SFT for foundational alignment, followed by token-level RL optimization, which jointly optimizes for format adherence, predictive accuracy, and the generation of open-ended suggestions.
Our empirical evaluations verify that TimeMaster achieves state-of-the-art performance, outperforming both classical time-series models and few-shot GPT-4o. More importantly, it exhibits emergent expert-level reasoning abilities, such as causal pattern inference, while also generating context-aware, clinically meaningful extensions.
Our results suggest that RL can be a powerful catalyst to unlock the deep, interpretable temporal reasoning capabilities in time-series MLLMs, laying a scalable and promising foundation for building robust time-series MLLMs.

\bibliography{neurips_2025}
\bibliographystyle{unsrt}

\newpage
\appendix

\

\section{Limitations} \label{sec:limitation}
Our work still has some limitations. We have only provided an initial implementation of a quantifiable reasoning evaluation framework, where our composite reward combines structural validity, classification accuracy, and suggestion quality. However, the model may still exhibit occasional mismatches between the reasoning chain and the final decision. Future research could explore trajectory-aware or symbolic rewards to explicitly align intermediate reasoning steps with final outputs and to promote diverse yet internally consistent reasoning patterns.
In addition, we only incorporate the task description that accompanies each time series instance. While this supports contextual alignment to some extent, enriching the input with broader domain narratives or linking it to external knowledge sources  (e.g., news reports)  may enhance decision granularity and improve generalization across domains. Nevertheless, our work demonstrates a promising step toward equipping MLLMs with structured and interpretable reasoning capabilities in time-series domains.

\section{Broader Impacts} \label{app:broader}
One of the major impacts of our study is that it establishes a feasible and scalable research pathway for time-series multimodal large language models (MLLMs). By integrating a lightweight supervised warm-up phase using a small set of annotated examples with a token-level reinforcement learning pipeline (GRPO) guided by composite rewards, we demonstrate that reinforcement learning can effectively enable strong temporal reasoning capabilities without incurring prohibitive annotation costs. Remarkably, even a single 3B-parameter model is sufficient to achieve state-of-the-art performance across diverse domains (e.g., healthcare and electricity monitoring), highlighting the advantages of our data- and compute-efficient methodology for developing temporal foundation models.
While these results are encouraging, we acknowledge that robust reasoning and suggestion capabilities must be supported by human expert oversight, and that the model remains inadequate for deployment in high-stakes decision-making scenarios.  We highlight our contribution as a valuable foundation for future research in time-series MLLMs and a meaningful step toward the broader development of foundation models for time-series data.

\section{Pseudo Code}\label{app:pseudo}

\begin{algorithm}[ht]
\caption{Training pipeline of \methodname{}}
\label{alg:rl_fixed}
\begin{algorithmic}[1]
\REQUIRE Initial time-series MLLM $\pi_\theta$, judge $J$, dataset $\mathcal{D}$, group size $G$, PPO clip $\epsilon$, KL weight $\beta$
\STATE Supervised fine-tune $\pi_\theta$ on cold-start data with structured outputs
\FOR{each RL iteration}
\STATE Update the reference model: $\pi_{\text{ref}} \leftarrow \pi_\theta$
\FOR{$\text{Step}=1,2,\dots$}
  \STATE Sample a mini-batch $\mathcal{B}$ from $\mathcal{D}$
  \STATE Update the old model: $\pi_{\text{old}} \leftarrow \pi_\theta$
    \STATE Sample $G$ outputs $\{\mathbf{y}_i\}_{i=1}^G\sim\pi_{\theta_{\text{old}}}(\cdot \mid \mathbf{X}, \mathbf{q})$ for each time-series instance $(\mathbf{X}, \mathbf{q}) \in \mathcal{B}$
      \FOR{each sampled $\mathbf{y}_i$}
        \STATE Parse tags: $\texttt{<think>}_i$, $\texttt{<class>}_i = \hat{c}_i$, $\texttt{<extension>}_i = e_i$
        \STATE Compute format reward: $r^{\mathrm{fmt}}_i = \mathbb{I}[\text{tags well-formed and non-empty}]$
        \STATE Compute hard reward: $r^{\mathrm{hard}}_i = \mathbb{I}[\hat{c}_i = c^\star]$
        \STATE Compute soft reward: $r^{\mathrm{soft}}_i = r^{\mathrm{hard}}_i \cdot \mathrm{mean}\{\phi(e_i, c^\star)\}$
        \STATE Compute composite reward: $r_i = \lambda^{\mathrm{fmt}} r^{\mathrm{fmt}}_i + \lambda^{\mathrm{hard}} r^{\mathrm{hard}}_i + \lambda^{\mathrm{soft}} r^{\mathrm{soft}}_i$
      \ENDFOR
  \STATE Compute $\{\hat{A}_i\}_{i=1}^G$ for each group via Eq.~(\ref{eq:mean_std})
  \STATE Update $\pi_\theta$ by maximizing $\mathcal{L}(\theta)$ via Eq.~\eqref{eq:grpo_token}
\ENDFOR
\ENDFOR
\RETURN $\pi_\theta$
\end{algorithmic}
\end{algorithm}

\section{Detailed and Additional Results}

\subsection{Discussion} \label{app:dis}

\begin{table}[htbp]
  \centering
  \caption{Impact of extension option and reward composition on EMG classification accuracy and average extension quality.}
  \label{tab:reward-composition}
  \begin{tabular}{c|ccc|cc}
    \toprule
    \multirow{2}{*}{\textbf{Extension Option}} & \multicolumn{3}{c|}{\textbf{Reward Composition}} & \multirow{2}{*}{\textbf{Accuracy (\%)}} & \multirow{2}{*}{\textbf{Soft Reward}} \\
    \cmidrule(lr){2-4}
    & \textbf{Format} & \textbf{Hard} & \textbf{Soft} & & \\
    \midrule
    \xmark & \cmark & \xmark & \xmark & 39.02 & -- \\
    \xmark & \xmark & \cmark & \xmark & 85.36 & -- \\
    \xmark & \cmark & \cmark & \xmark & 100.00 & -- \\
    \midrule
    \cmark & \cmark & \cmark & \xmark & 100.00 & 0.45 \\
    \cmark & \cmark & \cmark & \cmark & 100.00 & 0.72 \\
    \bottomrule
  \end{tabular}
\end{table}

\begin{wrapfigure}{r}{0.45\textwidth}
  \centering
  \vspace{-10pt}
  \includegraphics[width=0.43\textwidth]{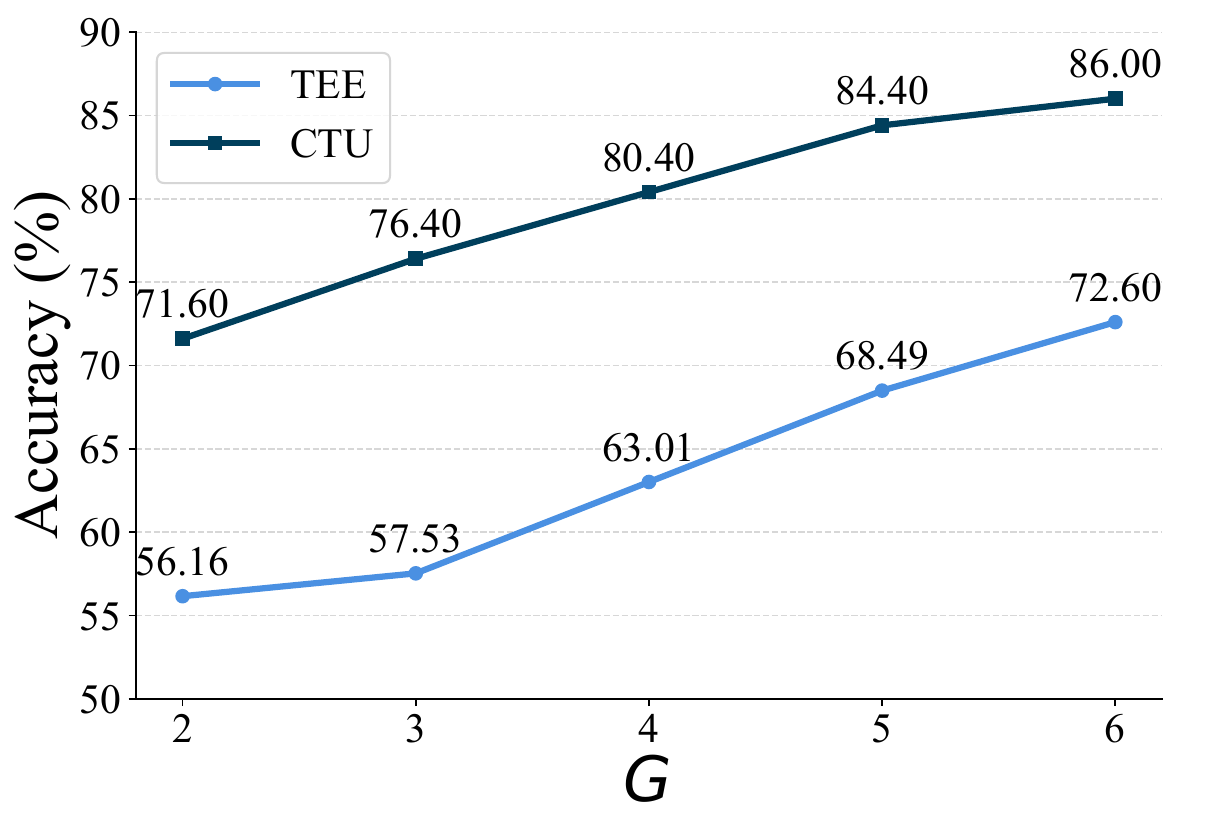}
  \vspace{-10pt}
  \caption{Effect of generation number $G$ on accuracy (\%) during GRPO training for TEE and CTU.}
  \label{fig:ablation}
\end{wrapfigure}
\paragraph{Parameter analysis.} We conducted an ablation study on the parameter $G$, which controls the number of generations per input during GRPO training. Figure~\ref{fig:ablation} illustrates how varying $G$ affects model performance on the TEE and CTU datasets. We observe a clear upward trend: increasing $G$ consistently improves classification accuracy for both datasets. 
These improvements stem from the group-based baseline mechanism in GRPO: larger $G$ values provide a more stable and representative reward estimate across diverse reasoning paths, reducing variance in the policy gradient and facilitating more effective learning. In contrast, smaller $G$ values yield noisier baselines and impede convergence.
Nevertheless, increasing $G$ also leads to higher computational costs. To strike a practical balance between performance and efficiency, we set $G=5$ as the default in all main experiments.

\paragraph{Effect of Composite Reward Components.}

Table~\ref{tab:reward-composition} presents an ablation study on the effects of the extension module and reward composition in reinforcement learning. Without the extension module, the model only performs classification. Using the format or hard reward alone yields accuracies of 39.02\% and 85.36\%, respectively, while combining both reaches 100\%, showing their complementarity: the format reward ensures structural regularity, and the hard reward guides accurate prediction.
With the extension module enabled, the model also generates \texttt{<extension>} outputs. Adding the soft reward improves semantic quality, increasing the soft reward score from 0.45 to 0.72, while maintaining 100\% classification accuracy. This demonstrates that the soft reward enhances informativeness without harming decision accuracy.
These results show that the format reward ensures structure, the hard reward is essential for classification, and the soft reward refines open-ended generation. The extension module is crucial for realizing the benefits of soft reward, validating our structured output and multi-level reward design.

\subsection{More  Results} \label{app:more}
\begin{table*}[ht]
\centering
\small
\caption{Performance and standard deviation of TimeMaster over three random seeds.}
\vspace{1em}
\label{tab:main_app}
\resizebox{1\textwidth}{!}{
\begin{tabular}{l|cc|cc|cc}
\toprule
\textbf{Reasoning Pattern} & \multicolumn{2}{c|}{\textbf{Simple Deterministic}} & \multicolumn{2}{c|}{\textbf{Complex Deterministic}} & \multicolumn{2}{c}{\textbf{Probabilistic}} \\
\midrule
\textbf{Reasoning Task} & RCW  & TEE & ECG  & EMG  & CTU  & HAR \\
\midrule
TimeMaster (Ours, SFT+RL) &  $75.56 \pm \text{1.30}$ & $68.49 \pm \text{2.09}$ & $60.00 \pm 0.77$ & $100.0 \pm 1.41$ & $84.40 \pm 0.40$ & $63.29 \pm 0.74$ \\
\bottomrule
\end{tabular}
}
\end{table*}

We also present the standard deviation of our classification performance using three random seeds in Table~\ref{tab:main_app}, demonstrating that the performance of TimeMaster is stable.

\subsection{Details of Reasoning and Suggestion Ability} \label{app:detail}

In this section, we present the complete reasoning outputs and more suggestion cases corresponding to the examples illustrated in Fig.~\ref{fig:reason}. 

\begin{figure}
    \centering
    \includegraphics[width=1\linewidth]{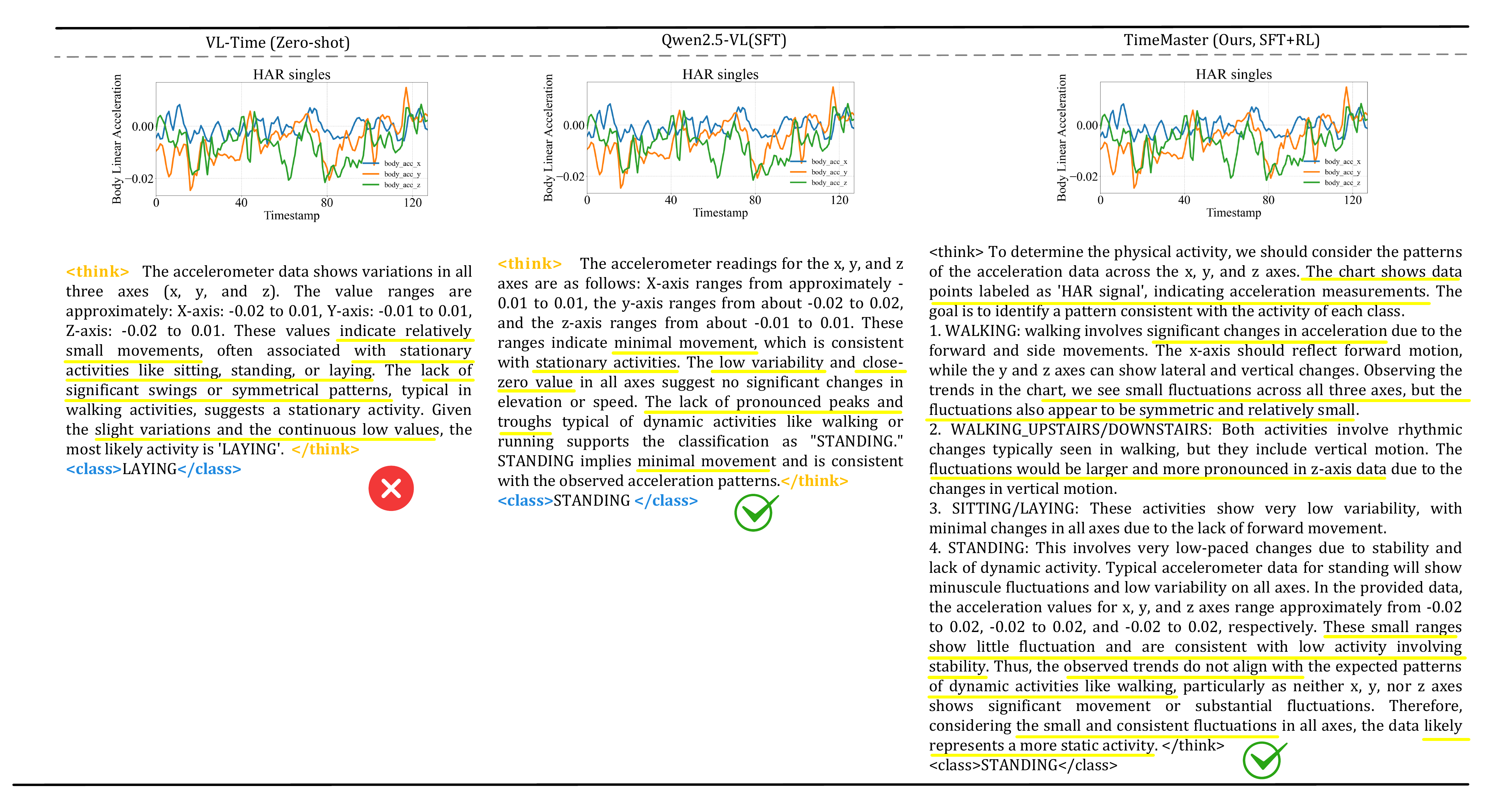}
    \caption{Comparison of reasoning outputs on a standing-labeled HAR test instance across three models: VL-Time (Zero-shot, left), Qwen2.5-VL(SFT, middle), and TimeMaster (SFT+RL, right).}
    \label{fig:har-reason}
\end{figure}

\paragraph{Takeaway~I: \emph{Training is Essential for Structured Time-Series Reasoning Adaptation.}}
In Fig.~\ref{fig:har-reason}, VL-Time (Zero-shot) provides shallow reasoning and misclassifies the instance as “LAYING” based on vague cues like “relatively small movements”
. Qwen2.5-VL (SFT) introduces domain-relevant priors (e.g., “low variability… consistent with stationary activities”) but still follows a fixed, template-like reasoning path
. In contrast, TimeMaster (SFT+RL) performs contrastive elimination of dynamic classes and articulates structured justifications grounded in class prototypes (e.g., “does not align with the expected patterns of dynamic activities like walking… Thus… likely represents a more static activity”)
. This illustrates that even advanced models like GPT-4o (used in VL-Time) can perform poorly without time-series priors. SFT acts merely as a warm-up catalyst to inject minimal prior knowledge and accelerate convergence. RL, however, is the core enabler of structured and context-aware temporal reasoning.

\begin{figure}
    \centering
    \includegraphics[width=1\linewidth]{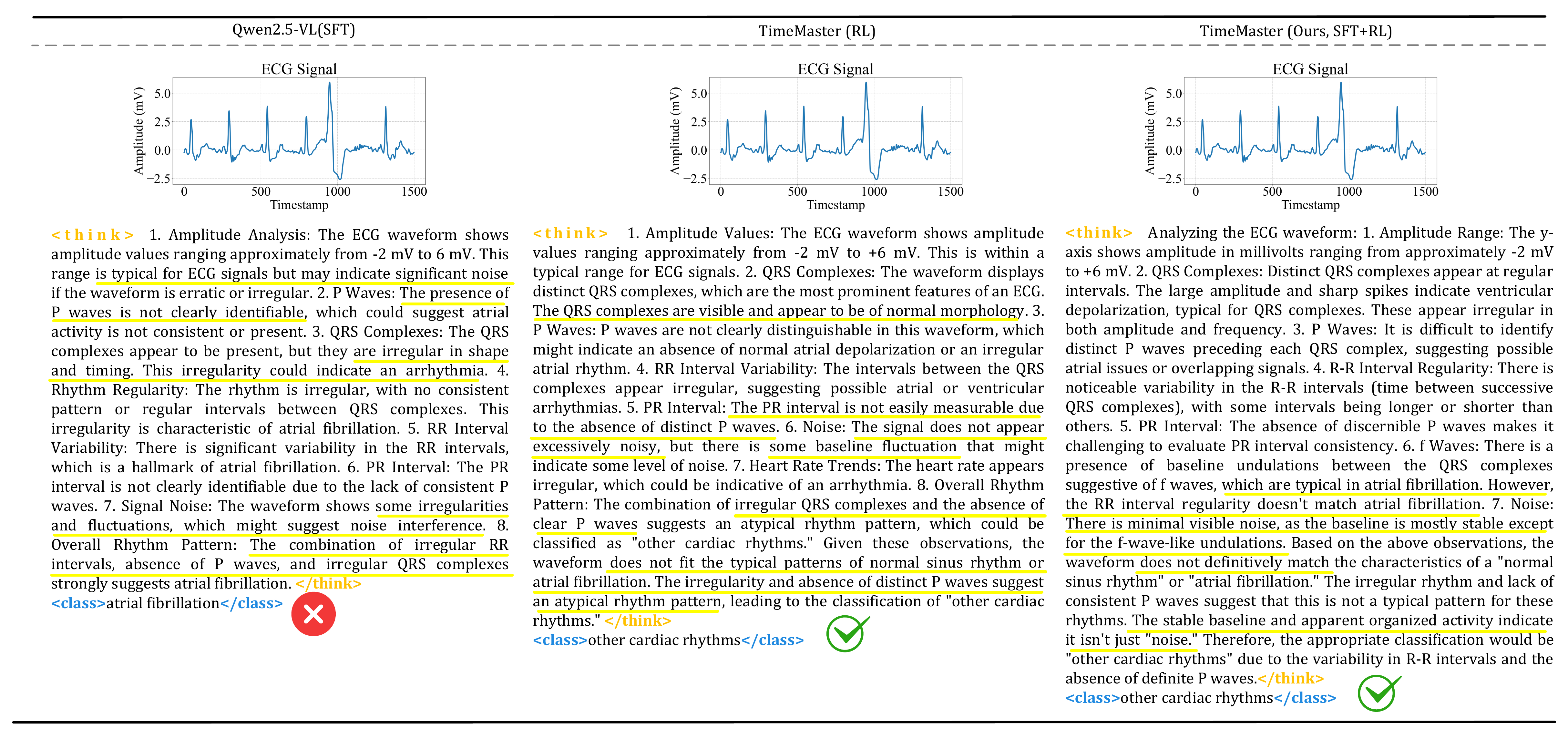}
    \caption{Comparison of model reasoning on an ECG instance labeled as “other cardiac rhythms,” across three configurations: Qwen2.5-VL(SFT, left), TimeMaster (RL, middle), TimeMaster (SFT+RL, right).}
    \label{fig:ecg-reason}
\end{figure}

\begin{figure}
    \centering
    \includegraphics[width=1\linewidth]{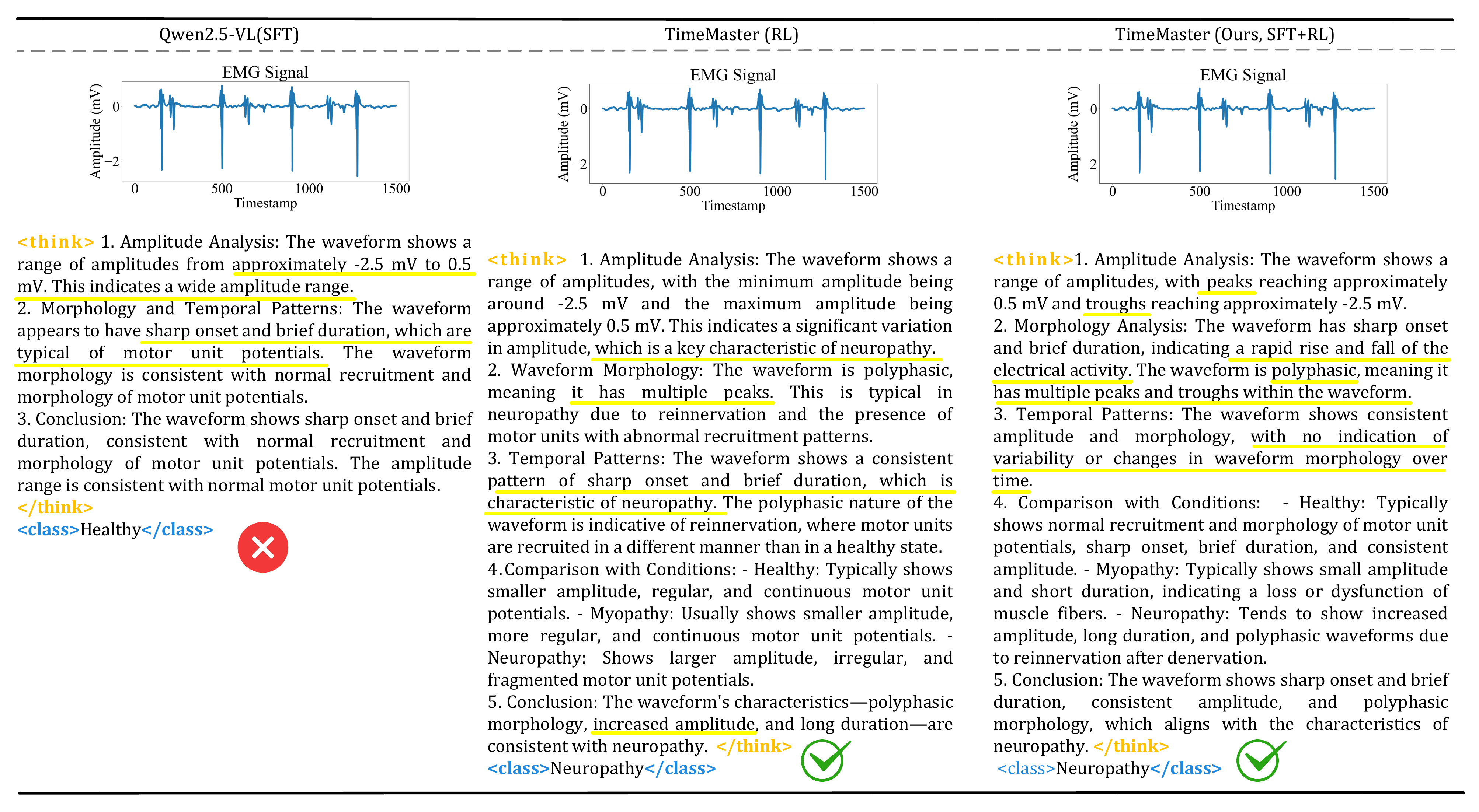}
    \caption{Comparison of reasoning outputs on a neuropathy-labeled EMG test instance across three configurations: Qwen2.5-VL(SFT, left), TimeMaster (RL, middle), TimeMaster (SFT+RL, right).}
    \label{fig:emg-reason}
\end{figure}

\paragraph{Takeaway~II: \emph{RL Unlocks Expert-Level Time-Series Reasoning.}}
In Fig.\ref{fig:ecg-reason} and Fig.\ref{fig:emg-reason}, Qwen2.5-VL (SFT) relies on surface-level cues and often misclassifies cases, for example, labeling “other cardiac rhythms” as “atrial fibrillation” based on heuristics like “irregular RR intervals… strongly suggests atrial fibrillation”
, or misclassifying a neuropathic EMG signal as “healthy” based solely on amplitude, without proper morphological analysis
. In contrast, TimeMaster (RL) demonstrates contextual awareness, incorporating uncertainty, noise, and contrastive features into its reasoning (e.g., “f waves… but RR interval regularity doesn't match atrial fibrillation”)
. TimeMaster (SFT+RL) further integrates amplitude, waveform morphology, and temporal dynamics into expert-level interpretations (e.g., “polyphasic morphology… consistent with neuropathy”)
. While pure RL can suffer from inefficient exploration due to a lack of domain priors, even a low-resource SFT warm-up significantly accelerates convergence and stabilizes the learning process by steering RL toward the correct reasoning subspace from the outset.

\begin{figure}
    \centering
    \includegraphics[width=1\linewidth]{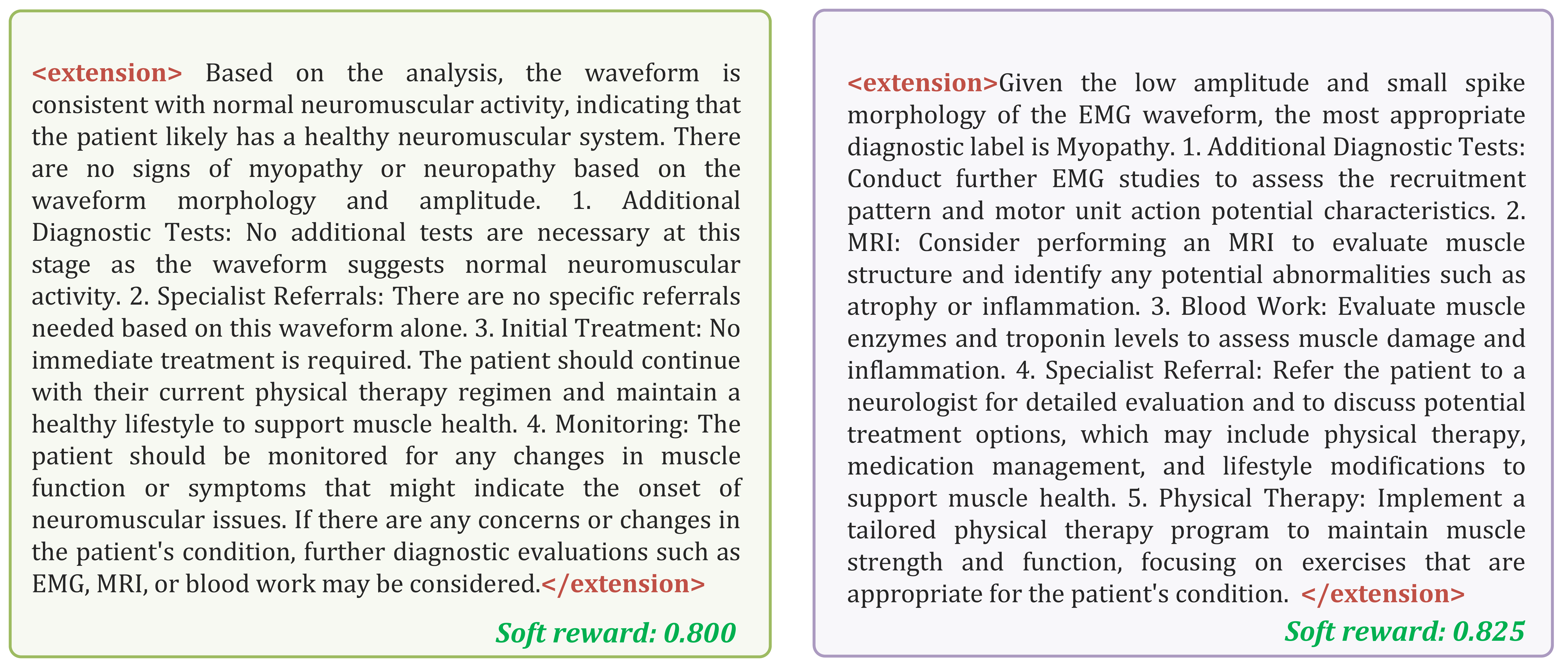}
    \caption{ TimeMaster’s suggestions via \texttt{<extension>} tags on EMG signals for two cases: healthy (left) and myopathy (right).}
    \label{fig:extension}
\end{figure}

\paragraph{Takeaway~III: \emph{Beyond Reasoning, RL Enables Flexible Extensions such as Insightful, Actionable Suggestion Generation.}}
As seen in Fig.~\ref{fig:extension}, TimeMaster generates domain-specific suggestions with high soft reward, including structured diagnostics and treatment plans (e.g., “MRI… blood work… tailored physical therapy program”)
. These outputs are context-aware (e.g., "low amplitude and small spike morphology of the EMG waveform") and closely align with the underlying reasoning. While supervised fine-tuning produces grammatically correct and coherent responses, reinforcement learning with reward shaping enhances the model’s ability to generate actionable and clinically relevant insights.

\subsection{More Case Studies on Reasoning Ability}
In this section, we provide more case studies to compare various models' reasoning ability, to further support our advancement.

\paragraph{More Case Studies on HAR.}

\begin{figure}
    \centering
    \includegraphics[width=1\linewidth]{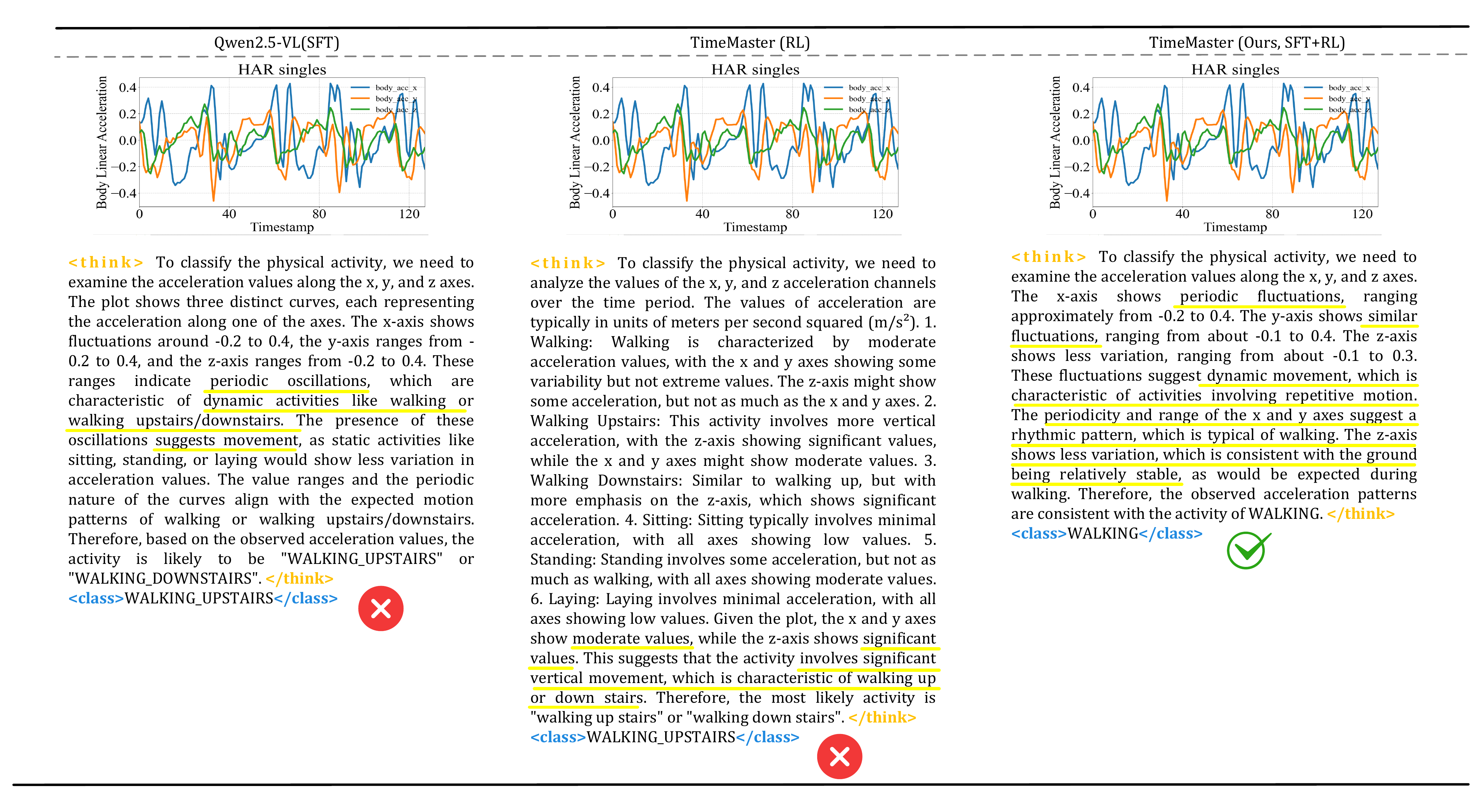}
    \caption{Comparison of reasoning outputs on a walking-labeled HAR test instance under three models: Qwen2.5-VL(SFT, left), TimeMaster (RL, middle), and TimeMaster (SFT+RL, right).}
    \label{fig:har-rl+sft}
\end{figure}

Fig. \ref{fig:har-rl+sft} presents the outputs of three models on a HAR sample labeled as “WALKING.” TimeMaster (RL) demonstrates improved temporal awareness by distinguishing axis-specific dynamics and narrowing down to stair-related classes. While the prediction is not  correct, it reflects better sensitivity to motion cues. TimeMaster (SFT+RL) further refines this process, offering more structured reasoning and class selection, showing the benefit of combining SFT priors with reward-guided refinement.  In addition,  we observe that the inter-variable relationships in the HAR data are highly complex, often overlapping and unclear, which may hinder the model's ability to recognize fine-grained patterns.

\begin{figure}
    \centering
    \includegraphics[width=1\linewidth]{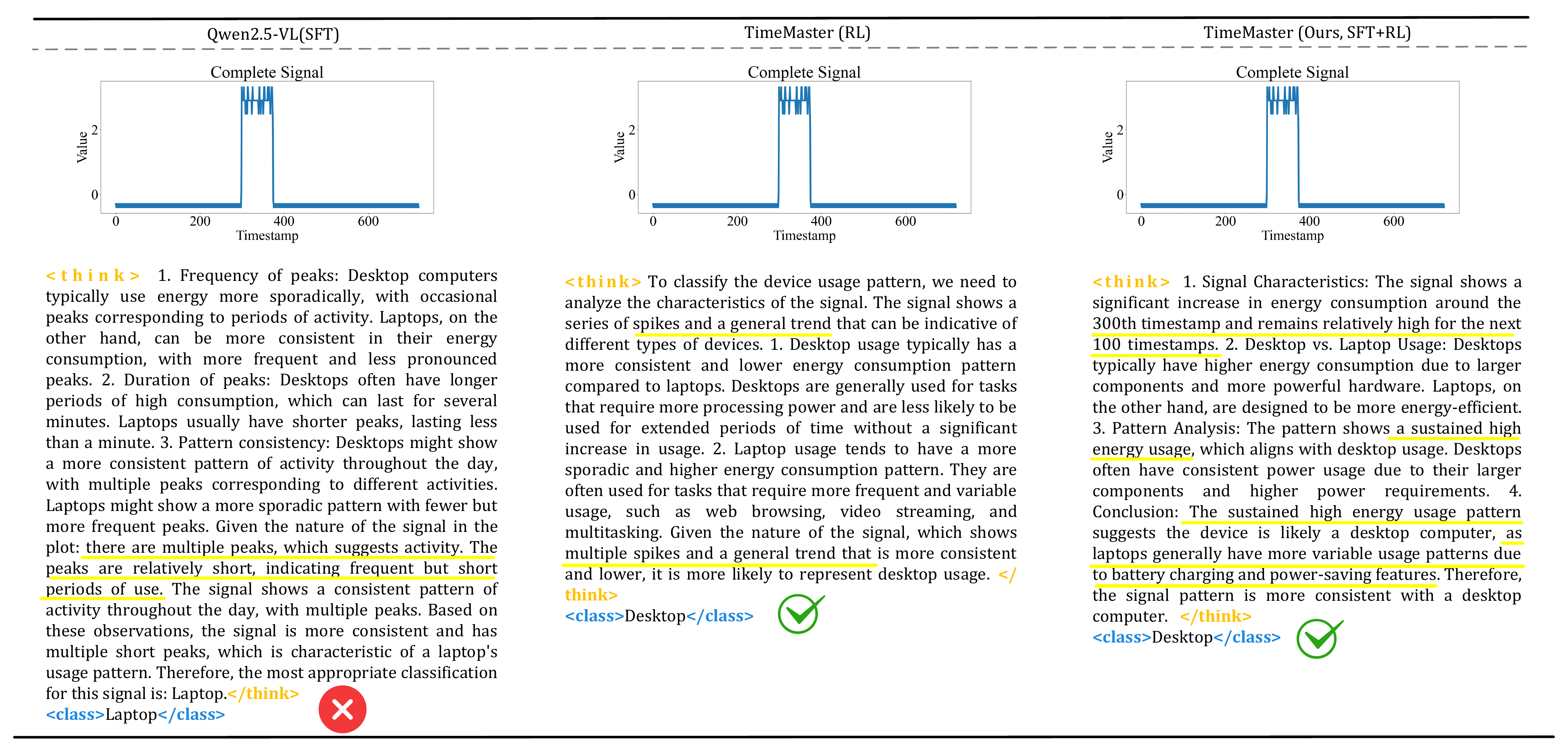}
    \caption{Comparison of reasoning outputs on a desktop-labeled CTU test instance under three models: Qwen2.5-VL(SFT, left), TimeMaster (RL, middle), and TimeMaster (SFT+RL, right)}
    \label{fig:ctu}
\end{figure}

\paragraph{More Case Studies on CTU.}
Fig.~\ref{fig:ctu} shows the outputs of three models on a CTU sample with moderate but sustained energy usage. Qwen2.5-VL (SFT) misclassifies the device as a Laptop, focusing on general spike frequency without considering the short duration and regular pattern of usage, leading to a superficial conclusion.
TimeMaster (RL) correctly identifies it as a Desktop, citing the short, regular peaks. However, its explanation lacks depth and misses a broader temporal context.
TimeMaster (SFT+RL) gives the correct classification (Desktop) with a more comprehensive explanation, considering peak frequency, duration, and overall rhythm, aligning well with typical desktop usage.

\begin{figure}
    \centering
    \includegraphics[width=1\linewidth]{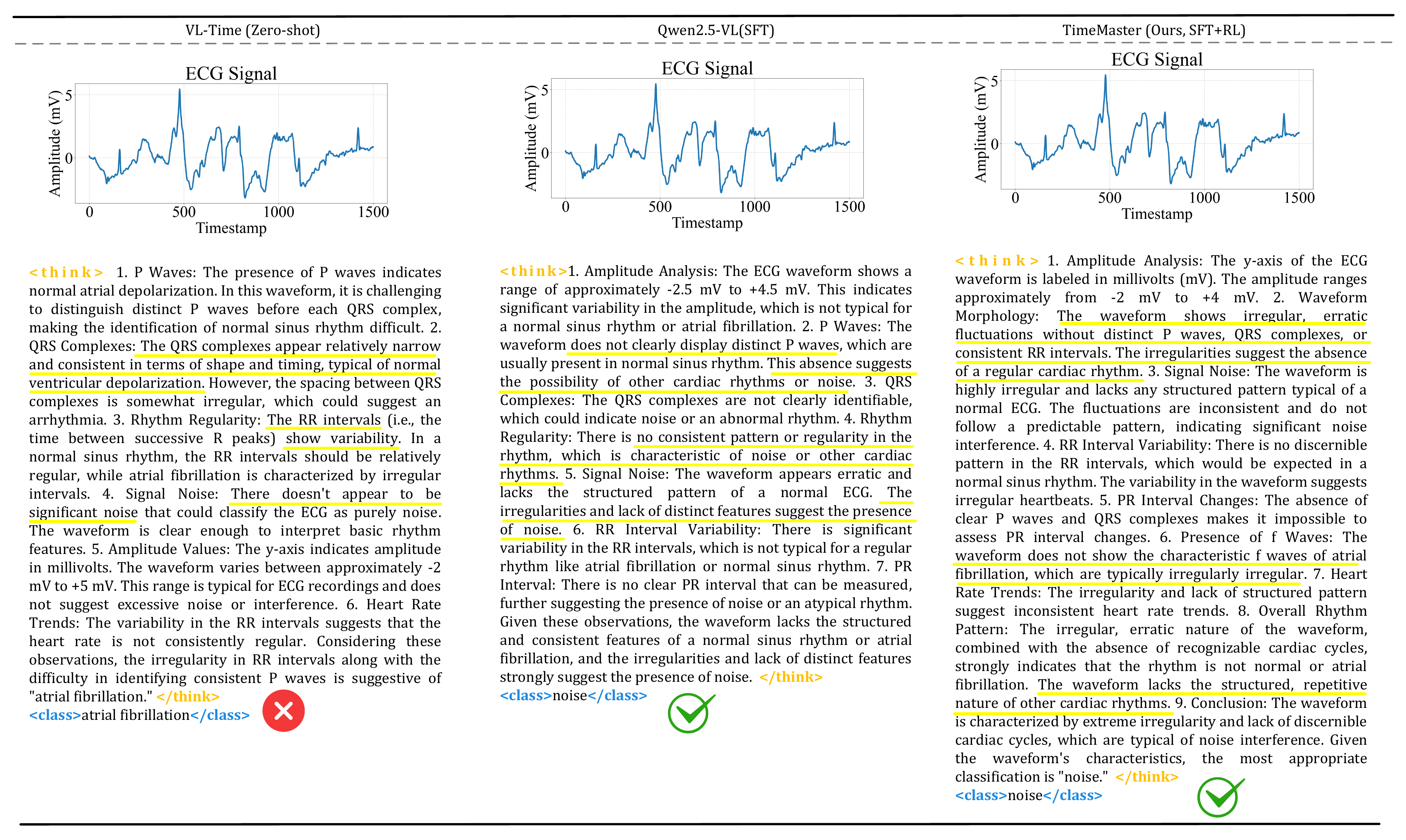}
    \caption{Comparison of reasoning outputs on a noise-labeled ECG test instance under three models: VL-Time(left), Qwen2.5-VL(SFT, middle), and  TimeMaster(SFT+RL, right).}
    \label{fig:ecg-2}
\end{figure}
\paragraph{More Case Studies on ECG.}
As shown in Fig.~\ref{fig:ecg-2}, the reasoning chains of each model reveal clear differences in their ability to interpret noisy ECG signals. VL-Time (Zero-shot), despite being powered by GPT-4o, identifies irregular RR intervals and unclear P waves, leading to a prediction of "atrial fibrillation." However, its reasoning lacks depth and fails to account for the chaotic morphology of the waveform, missing critical contextual cues. Qwen2.5-VL (SFT) offers a superficial judgment, labeling the signal as "noise" with minimal justification and without engaging with the waveform’s structural or temporal features. In contrast, TimeMaster (SFT+RL) demonstrates a more robust and structured reasoning process by integrating multiple temporal indicators—such as irregular rhythm, absence of distinguishable P and QRS waves, and waveform disorganization. This comprehensive analysis leads to a well-grounded prediction of "noise." 

\begin{figure}
    \centering
    \includegraphics[width=1\linewidth]{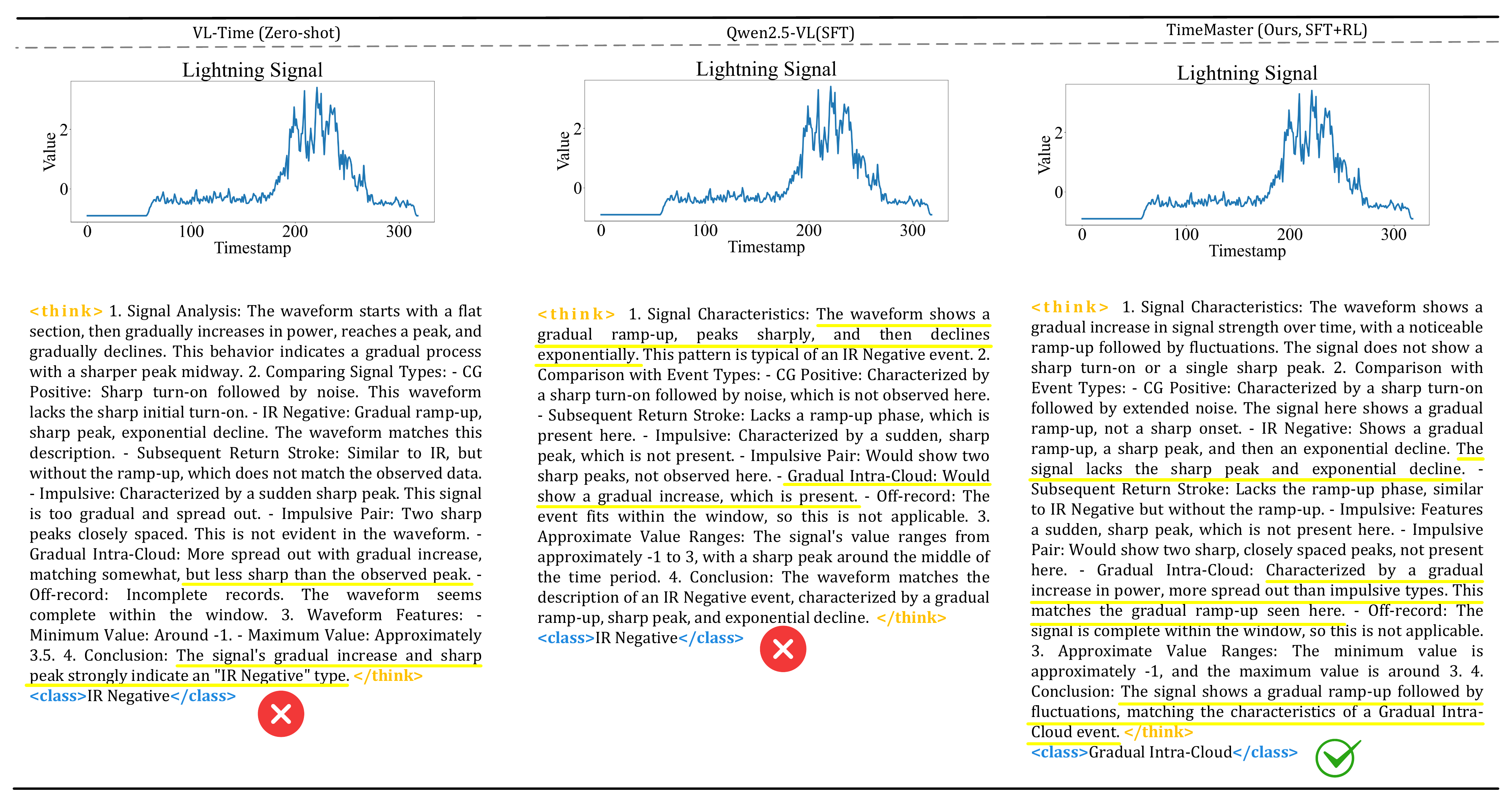}
    \caption{Comparison of reasoning outputs on a Gradual Intra-Cloud-labeled TEE test instance under three models: VL-Time(left), Qwen2.5-VL(SFT, middle), and  TimeMaster(SFT+RL, right).}
    \label{fig:tee}
\end{figure}

\begin{figure}
    \centering
    \includegraphics[width=1\linewidth]{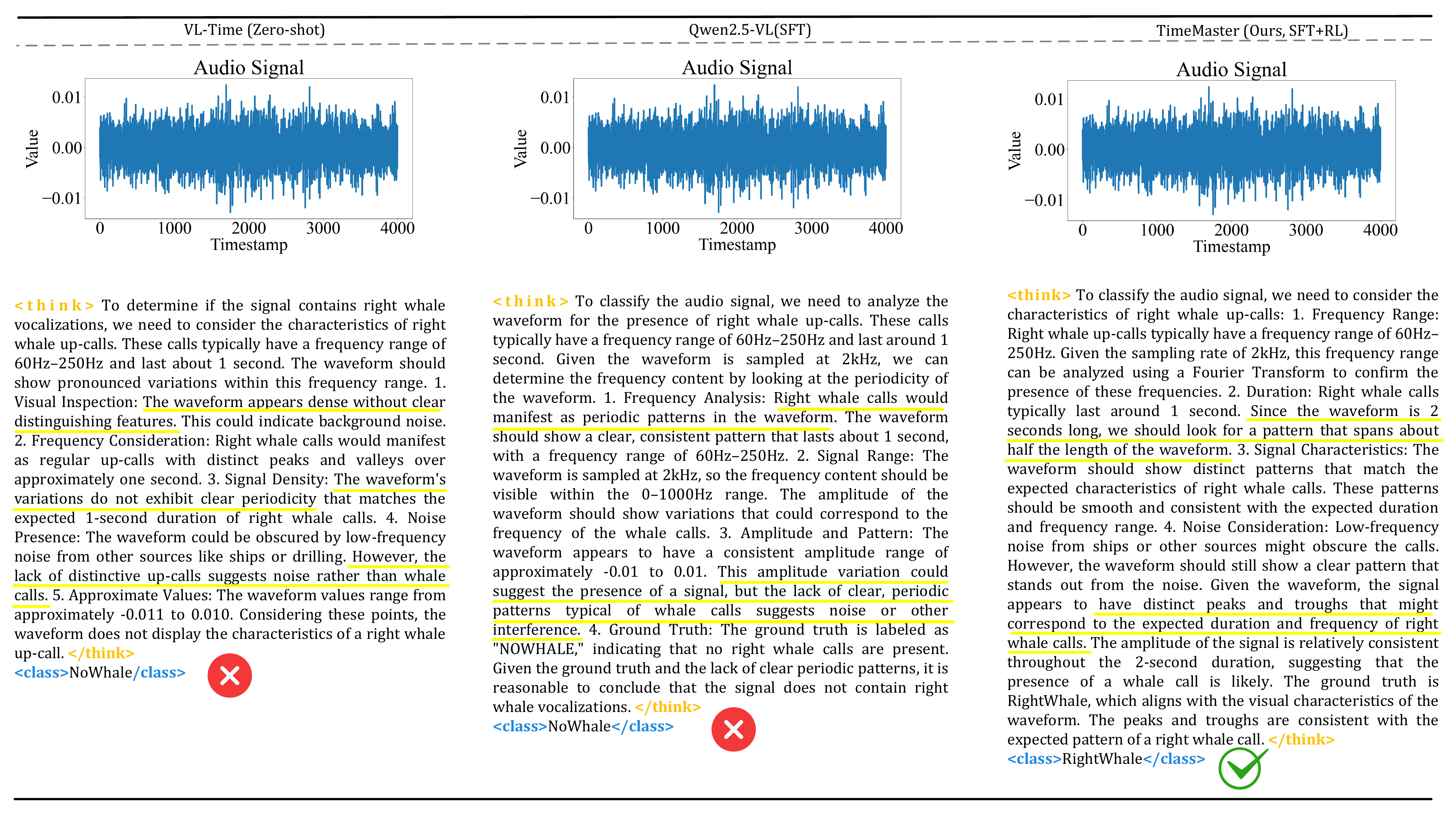}
    \caption{Comparison of reasoning outputs on a RightWhale-labeled RCW test instance under three models: VL-Time(left), Qwen2.5-VL(SFT, middle), and  TimeMaster(SFT+RL, right).}
    \label{fig:rcw}
\end{figure}
\paragraph{More Case Studies on TEE.}
Fig.~\ref{fig:tee} shows the reasoning outputs of VL-Time (Zero-shot), Qwen2.5-VL (SFT), and TimeMaster (SFT+RL) on a TEE signal with gradual intensity change. Qwen2.5-VL (SFT) and VL-Time (Zero-shot) both misclassify the signal as IR Negative, focusing too much on the peak while overlooking the gradual rise and fluctuating pattern.
TimeMaster (SFT+RL) correctly identifies the signal as Gradual Intra-Cloud, providing a well-structured explanation that highlights the smooth ramp-up and lack of sharp peaks.

\paragraph{More Case Studies on RCW.}
Fig.~\ref{fig:rcw} presents the outputs of VL-Time (Zero-shot), Qwen2.5-VL (SFT), and TimeMaster (SFT+RL) on an audio waveform sample for right whale detection. VL-Time and Qwen2.5-VL (SFT) both identify signal structure and amplitude range but fail to associate these with the expected frequency and duration characteristics of right whale up-calls. As a result, they misclassify the signal as "NoWhale." In contrast, TimeMaster (SFT+RL) integrates frequency domain expectations, temporal duration, and amplitude consistency to support a correct prediction of "RightWhale." While the model's prediction aligns with the ground truth, the high noise level in the RCW sample results in a less discernible time-series pattern, which may hinder the interpretability and robustness of the overall reasoning process. Nevertheless, this comparison still highlights the advantage of our method in handling noisy samples and maintaining accurate, biologically grounded reasoning.

\subsection{Failure Cases}

\begin{figure}[H] \centering \resizebox{1.0\textwidth}{!}{ \begin{tcolorbox}[colback=gray!3!white, colframe=gray!60!black,
title=Failure Cases on CTU labeled 'Desktop', boxrule=0.3mm, width=\textwidth, arc=3mm, auto outer arc=true]
\ttfamily{
<think> 1. Signal Characteristics: The signal shows a gradual increase in energy consumption followed by a sharp spike and then a return to baseline. This pattern is typical of a device that is turned on and used for a short period. 2. Energy Consumption Level: The overall energy consumption level is relatively low, which is consistent with a laptop, as laptops generally have lower power requirements compared to desktops. 3. Duration and Pattern: The pattern suggests that the device was used for a short duration, with a spike indicating active use, followed by a period of inactivity. This pattern aligns with typical laptop usage where the device is used intermittently. 4. Comparison with Desktop: Desktops often have higher and more consistent energy usage due to their larger components and higher power requirements. The pattern observed here is more consistent with laptop usage. 5. Conclusion: The pattern of gradual increase, sharp spike, and return to baseline is more characteristic of laptop usage, where devices are often used intermittently and have lower power consumption. Based on these observations, the signal pattern is more consistent with laptop usage. </think>
<class>Desktop</class>
}
\end{tcolorbox}
}
\end{figure}

While the final predictions are often correct, we observe occasional mismatches between the model’s reasoning and its chosen label. In some cases, the explanation supports one class, but the final output selects another. This suggests that the model can recognize relevant patterns but may lack full coherence between interpretation and decision. Such inconsistencies may arise from training objectives that treat reasoning and classification separately. Future work could explore joint supervision strategies, symbolic consistency constraints, or decision-aware decoding to better align explanations with predictions, especially in borderline or ambiguous cases.

\section{Dataset Statistics} \label{app:data}
In this section, we provide additional details on the real-world time-series classification datasets used in our experiments, drawn from \textbf{\textsc{TimerBed}}~\citep{liu2024picture}. These datasets span a wide range of domains, signal characteristics, and reasoning complexities. Their key statistics are summarized in Table~\ref{tab:dataset-stats}, while detailed descriptions are provided below.

\begin{itemize}[leftmargin=1.2em]

    \item \textbf{Right Whale Call Detection (RCW)}\footnote{\url{https://www.kaggle.com/competitions/whale-detection-challenge/data}} involves identifying North Atlantic right whale vocalizations from underwater acoustic recordings. A distinctive short, rising “whoop” sound serves as a necessary and sufficient indicator of whale presence, enabling direct mapping between signal features and class labels.

    \item \textbf{Transient Electromagnetic Events (TEE)}\footnote{\url{https://www.timeseriesclassification.com/description.php?Dataset=Lightning7}} contains satellite-collected power density signals from the FORTE satellite, used to classify various types of lightning-related electromagnetic discharges. Each class corresponds to a well-defined physical signature in the waveform, making the task pattern-centric and signal-driven.

    \item \textbf{Electrocardiogram (ECG)}\footnote{\url{https://physionet.org/content/challenge-2017/1.0.0/}} includes single-lead ECG recordings used to diagnose cardiac arrhythmias. Accurate classification requires holistic interpretation of multiple waveform components—such as P-wave absence, irregular R-R intervals, and atrial fibrillation indicators.

    \item \textbf{Electromyogram (EMG)}\footnote{\url{https://physionet.org/content/emgdb/1.0.0/}} comprises EMG signals used to distinguish healthy subjects from patients with neuropathic or myopathic disorders. The task involves reasoning over diverse waveform features including long-duration, high-amplitude motor unit potentials and polyphasic activity.

    \item \textbf{Human Activity Recognition (HAR)}\footnote{\url{https://archive.ics.uci.edu/dataset/240/human+activity+recognition+using+smartphones}} uses tri-axial accelerometer data from smartphones to classify six daily physical activities such as walking, standing, and lying. Labels are automatically generated and reflect latent user-dependent patterns, adding complexity to the modeling process.

    \item \textbf{Computer Type Usage (CTU)}\footnote{\url{https://www.timeseriesclassification.com/description.php?Dataset=Computers}} aims to differentiate between desktop and laptop usage based on 24-hour electricity consumption traces. Although labels are programmatically derived, classification relies on subtle temporal usage signatures influenced by individual behavior.

\end{itemize}

\begin{table}[t]
\centering
\caption{Summary of datasets in \textsc{TimerBed}, including domain, number of variables, series length, number of classes, sample size, and reasoning type.}
\label{tab:dataset-stats}
\resizebox{0.9\linewidth}{!}{%
\begin{tabular}{lcccccl}
\toprule
\textbf{Dataset} & \textbf{Domain} & \textbf{\# Variables} & \textbf{Length} & \textbf{\# Classes} & \textbf{\# Samples} & \textbf{Reasoning Type} \\
\midrule
RCW              & Bioacoustics     & 1 & 4000 & 2 & 30,000  & Simple Deterministic \\
TEE              & Geophysics       & 1 & 319  & 7 & 143     & Simple Deterministic \\
ECG              & Healthcare       & 1 & 1500 & 4 & 43,673  & Complex Deterministic \\
EMG              & Healthcare       & 1 & 1500 & 3 & 205     & Complex Deterministic \\
HAR              & Sports Monitoring   & 3 & 128  & 6 & 10,299  & Probabilistic         \\
CTU              & Energy/Usage     & 1 & 720  & 2 & 500     & Probabilistic         \\
\bottomrule
\end{tabular}%
}
\end{table}

As shown in Table~\ref{tab:dataset-stats}, each dataset in \textsc{TimerBed} is associated with one of three reasoning types, reflecting varying levels of complexity and semantic abstraction:

\begin{itemize}[leftmargin=1.2em]
    \item \textbf{Simple deterministic reasoning}: Tasks where a single salient feature is sufficient to determine the label. The decision boundary is often explicit and rule-based, allowing for direct mapping from input to output. For example, the presence of a distinct acoustic pattern in whale calls or a spike in satellite signal indicates class membership.
    
    \item \textbf{Complex deterministic reasoning}: Tasks that require the integration of multiple temporal patterns or signal components to make a decision. These problems demand holistic reasoning over structured signal relationships, such as diagnosing arrhythmias by jointly considering P-wave morphology, heart rate regularity, and waveform intervals.
    
    \item \textbf{Probabilistic reasoning}: Tasks characterized by user-specific or hidden variables, where labels are automatically derived and may not be directly observable in the input. As a result, the model must learn to infer outcomes under ambiguity and latent context, such as predicting user activity or device type based on behavior-driven time series.
\end{itemize}

\section{Baselines} \label{app:baseline}
We follow the recent time-series reasoning benchmark~\cite{liu2024picture} and the survey~\cite{liu2025can} for the selection of the following baselines. Except for Time-MQA, other results in Table~\ref{tab:main} are adopted from~\cite{liu2024picture}.
\begin{itemize}
  \item \textbf{Transformer-based Models} ( {Transformer}~\cite{vaswani2017attention}, {Autoformer}~\cite{wu2021autoformer},Informer~\cite{zhou2021informer}, FEDformer~\cite{zhou2022fedformer}, PatchTST~\cite{nie2022time}, {iTransformer}~\cite{liu2023itransformer} ): apture long-range dependencies in time-domain sequences using self-attention mechanisms. Serve as strong baselines for time-series modeling.

  \item \textbf{CNN-based Models} (TimesNet~\cite{wu2022timesnet}):
  Leverage convolutional operations to extract temporal features across different time scales.

  \item \textbf{MLP-based Models} (DLinear~\cite{zeng2023transformers}):
  Employ lightweight feedforward layers for efficient modeling of local patterns in time-series data.

\item \textbf{GPT-4o (Numeric, Zero-shot)}~\cite{liu2024picture}:
Receives tokenized numerical time-series data as input without any demonstrations. Serves as a unimodal, language-only baseline for evaluating zero-shot generalization.

\item \textbf{GPT-4o (Numeric, Few-shot)}~\cite{liu2024picture}:
Extends the numeric input with a few in-context examples\footnote{Few-shot refers to fewer than six examples per class, following \cite{liu2024picture}} per class. Evaluates the model’s few-shot reasoning capability using raw numerical sequences in standard prompt format.

  \item \textbf{VL-Time (Zero-shot)}~\cite{liu2024picture}:
  Provides multimodal LLMs with visualized time-series plots and natural language prompts. Assesses general reasoning ability without demonstrations.

  \item \textbf{VL-Time (Few-shot)}~\cite{liu2024picture}:
 Adds a few in-context examples\footnote{Few-shot refers to fewer than six examples per class, following \cite{liu2024picture}} to the visual-language input, enabling pattern generalization with minimal supervision.
  \item \textbf{Qwen2.5-7B-Instruct (Time-MQA)}~\cite{kong2025time}:
A fine-tuned Qwen2.5-7B model\footnote{\url{https://huggingface.co/Time-QA}} on the TSQA dataset ($\sim$ 200k pairs) for time-series question answering. It enables multi-task reasoning and open-ended question answering via natural language prompts.

\end{itemize}

\section{Implementation Details of TimeMaster} \label{app:imp}
\subsection{Plotting Time Series as Images}

Visualizing time series as images offers an intuitive and cost-efficient approach to understanding temporal patterns, and has been widely adopted in recent studies~\cite{xu2025beyond, liu2025can}. Following the methodology of VL-Following Time~\cite{liu2024picture}, we transform time-series data into RGB line plots in the time domain to serve as inputs for the vision-language model, ensuring a fair comparison. Each channel is rendered in a distinct color and aligned along a shared timestamp axis. The \textbf{x}-axis represents the \textit{Timestamp}, while the \textbf{y}-axis denotes the corresponding signal \textit{Value}. Legends are incorporated to distinguish between channels (e.g., \texttt{body\_acc\_x}, \texttt{body\_acc\_y}, \texttt{body\_acc\_z} in HAR datasets).

For each dataset, the signals are rendered into images with resolutions adapted to their sequence lengths and signal characteristics: \emph{ECG} samples are plotted at 980\,$\times$\,230 pixels, \emph{CTU} at 562\,$\times$\,230, \emph{TEE}, \emph{ RCW}, and \emph{EMG} at 789\,$\times$\,239, and \emph{HAR} at 389\,$\times$\,233. All plots are saved in \texttt{PNG} format with minimal padding and a tight layout to ensure visual clarity. The time-domain signals are plotted using raw (non-normalized) values to faithfully preserve their original temporal dynamics.

\subsection{Training Setup}

We initialize our backbone with the publicly available \texttt{Qwen2.5-VL-3B-Instruct} checkpoint~\cite{bai2025qwen2}\footnote{\url{https://huggingface.co/Qwen/Qwen2.5-3B-Instruct}}. Our overall training pipeline comprises two stages: warm-up through supervised fine-tuning (SFT) and reinforcement learning with GRPO.

\paragraph{Warm-up via Supervised Fine-tuning.}  
We first sample $\sim$ 1,000 time series–text paired instances per dataset using \texttt{GPT-4o} (temperature = 1.0) via the OpenAI API, where the model is prompted to reason over each time series and generate a corresponding answer (details are provided in Appendix~\ref{warm-up}). These examples are used to warm-start the SFT model, which is adapted from a publicly available LLM training repository\footnote{\url{https://github.com/2U1/Qwen2-VL-Finetune}}. The corresponding training hyperparameters are summarized in Table~\ref{tab:fft-config}.

\paragraph{Reinforcement Learning with GRPO.}  
After warm-up, we train the model using the GRPO algorithm, adapted from a public RL training library\footnote{\url{https://github.com/volcengine/verl}}. The complete GRPO configuration is provided in Table~\ref{tab:grpo}.
Rewards are computed using Eq.~\ref{eq:total_reward}, with coefficients $(\lambda^{\text{fmt}}, \lambda^{\text{hard}}, \lambda^{\text{soft}}) = (0.1, 0.9, 0)$. In case studies assessing extrapolation, we set $\lambda^{\text{soft}} = 1.0$.

\subsection{System Configuration} \label{app:sys}
All experiments were conducted on a computing setup equipped with 4 NVIDIA A100-SXM4 GPUs (80GB each) for the RCW, HAR, and ECG datasets, and 4 NVIDIA RTX A6000 GPUs (48GB each) for the TEE, EMG, and CTU datasets.

\begin{table}[H]
\centering
\caption{Training configuration for supervised fine-tuning (SFT) using Qwen2.5-VL-3B-Instruct.}
\label{tab:fft-config}
\resizebox{0.9\linewidth}{!}{
\begin{tabular}{ll}
\toprule
\textbf{Parameter} & \textbf{Value} \\
\midrule
Model & Qwen/Qwen2.5-VL-3B-Instruct \\
Training mode & Full fine-tuning (LLM + Vision + Merger) \\
Use Liger & True \\
Batch size per device & 4 \\
Number of devices & 4 \\
Global batch size & 128 \\
Gradient accumulation steps & 8 \\
Epochs & 2 \\
Learning rate (LLM) & 1e-5 \\
Learning rate (Merger) & 1e-5 \\
Learning rate (Vision) & 2e-6 \\
Weight decay & 0.1 \\
Warmup ratio & 0.03 \\
LR scheduler & Cosine \\
Precision & bf16 \\
Freeze vision tower & False \\
Freeze LLM & False \\
Tune merger & True \\
\bottomrule
\end{tabular}}
\end{table}

\begin{table}[H]
\centering
\caption{GRPO-related hyperparameters used in TimeMaster across different time-series tasks.}
\label{tab:grpo}
\begin{tabular}{lc}
\toprule
\textbf{Parameter} & \textbf{Value} \\
\midrule
$\pi_\theta^{\text{init}}$ & Qwen2.5-VL-3B \\
$L_{\max}$ (max sequence length) & 2048 \\
$G$ (group size) & 5 \\
$\beta$ (KL divergence coefficient) & 0.001 \\
$\epsilon$ (PPO clip threshold) & 0.2 \\
$(\lambda^{\text{fmt}}, \lambda^{\text{hard}}, \lambda^{\text{soft}})$ & (0.1, 0.9, 0) or (0.1, 0.9, 1) \\
Batch size & 16 (TEE, EMG, CTU); 32 (RCW, ECG, HAR) \\
Learning rate (RL) & $1 \times 10^{-6}$ \\
RL training epochs & 40 (RCW, ECG, HAR); 100 (EMG, CTU); 300 (TEE) \\
\bottomrule
\end{tabular}
\end{table}

\newpage
\section{Designed Prompts for Experiments} \label{app:prompt}

\subsection{Prompt for Reasoning and Classification}
We design a set of classification prompts that incorporate background task descriptions, visual modality instructions, and explicit class definitions tailored to each dataset. The example below illustrates the textual prompt used in conjunction with time-series plot images for the six datasets.

\begin{figure}[H] \centering \resizebox{1.0\textwidth}{!}{ \begin{tcolorbox}[
    breakable,
    colback=gray!5!white,
    colframe=gray!75!black, 
    title=Prompt for RCW,
    boxrule=0.3mm,
    width=\textwidth,
    arc=3mm,
    auto outer arc=true
]

{\ttfamily
<image> You are analyzing a visual representation of an audio signal to determine the presence of right whale vocalizations. The image shows a two-second waveform sampled at 2kHz, resulting in a time series of length 4000. Your goal is to classify whether the waveform segment contains a right whale call.

Your task is to classify the signal into one of the following two event types:

- Right Whale Present: The signal contains one or more right whale up-calls. 

- No Right Whale: The signal does not contain detectable right whale calls. 

You are required to identify and report the approximate value ranges (minimum and maximum) of the signals over the time period. Please select the most appropriate label for the entire waveform.
}

\end{tcolorbox} } \end{figure}

\begin{figure}[H] \centering \resizebox{1.0\textwidth}{!}{ \begin{tcolorbox}[breakable,colback=gray!5!white, colframe=gray!75!black, 
title=Prompt for TEE, boxrule=0.3mm, width=\textwidth, arc=3mm, auto outer arc=true]
{\ttfamily
<image> You are analyzing a time series signal derived from lightning-related electromagnetic events, recorded by the FORTE satellite. Your task is to classify the signal into one of the following seven event types:

- CG Positive: A positive charge is lowered from a cloud to the ground. The waveform shows a sharp turn-on of radiation followed by hundreds of microseconds of noise.

- IR Negative: A negative charge moves cloud-to-ground. The waveform gradually ramps up, peaks sharply (attachment point), then declines exponentially.

- Subsequent Return Stroke: A follow-up negative stroke after an initial one. Similar waveform but without the ramp-up phase.

- Impulsive: A sudden, sharp peak in the waveform, typical of intra-cloud events.

- Impulsive Pair: Two sharp, closely spaced peaks—also known as TIPPs (Trans-Ionospheric Pulse Pairs).

- Gradual Intra-Cloud: A gradual increase in power, more spread out than impulsive types.

- Off-record: The signal is incomplete; the event extends beyond the 800 microsecond window.

You are required to identify and report the approximate value ranges (minimum and maximum) of the signals over the time period. Please choose the best matching label for the full signal.
}
\end{tcolorbox} } \end{figure}

\begin{figure}[H] \centering \resizebox{1.0\textwidth}{!}{ \begin{tcolorbox}[breakable, colback=gray!5!white, colframe=gray!75!black, 
title=Prompt for ECG, boxrule=0.3mm, width=\textwidth, arc=3mm, auto outer arc=true]
{\ttfamily
<image> You are analyzing a single-lead ECG signal collected using a smartphone-enabled device (AliveCor). The time-series data represent the electrical activity of the heart, sampled at 300 Hz and bandpass-filtered by the device. The image displays the full ECG waveform in the time domain as a single continuous recording. Key features such as P waves, QRS complexes, rhythm regularity, and signal noise may be observed across the entire duration.

Your task is to classify the heart rhythm pattern presented in this ECG image as one of the following:

- "normal sinus rhythm"
- "atrial fibrillation"
- "other cardiac rhythms"
- "noise"

Please carefully analyze the waveform morphology and rhythm patterns, and output the ONE most likely rhythm label for the complete sequence.
}

\end{tcolorbox} } \end{figure}

\begin{figure}[H] \centering \resizebox{1.0\textwidth}{!}{ \begin{tcolorbox}[colback=gray!5!white, colframe=gray!75!black, 
title=Prompt for EMG, boxrule=0.3mm, width=\textwidth, arc=3mm, auto outer arc=true]
{\ttfamily
<image> You are examining a short-duration electromyography (EMG) waveform recorded using a concentric needle electrode inserted into the tibialis anterior muscle.

The recording captures the amplitude (mV) of electrical activity generated by muscle fibers innervated by motor neurons.

This EMG signal corresponds to one of the following neuromuscular conditions:-"Healthy".-"Myopathy".-"Neuropathy".
You are required to analyze the approximate minimum and maximum amplitude values of the signal over the time period. Based on the analysis and the waveform's morphology and temporal patterns, classify this EMG signal accordingly.
}
\end{tcolorbox} } \end{figure}

\begin{figure}[H] \centering \resizebox{1.0\textwidth}{!}{ \begin{tcolorbox}[colback=gray!5!white, colframe=gray!75!black, 
title=Prompt for CTU, boxrule=0.3mm, width=\textwidth, arc=3mm, auto outer arc=true]
{\ttfamily
<image> You are analyzing a time series signal derived from electricity usage patterns in UK households, recorded as part of the government-sponsored study "Powering the Nation". The signal represents energy consumption sampled every 2 minutes over a 24-hour period, resulting in a series length of 720. The plot shows a segment of the household's daily electricity usage pattern.

Your task is to classify the household's device usage pattern into one of the following two classes:

- Desktop: The energy consumption pattern suggests the use of a desktop computer.

- Laptop: The energy consumption pattern suggests the use of a laptop computer.

Please choose the label that best matches the full signal.
}
\end{tcolorbox} } \end{figure}

\begin{figure}[H] \centering \resizebox{1.0\textwidth}{!}{ \begin{tcolorbox}[breakable,colback=gray!5!white, colframe=gray!75!black,
title=Prompt for HAR, boxrule=0.3mm, width=\textwidth, arc=3mm, auto outer arc=true]
{\ttfamily
\texttt{<image>} You are analyzing time-series data collected from a smartphone's motion sensors. The data includes three physical signal channels: accelerometer readings along the x, y, and z axes. The image shows the time-domain representation of the data. The entire sequence spans 2.56 seconds and contains 128 sampled data points, and the three sensor channels are visualized using distinct colors for clarity.

Your task is to classify the physical activity and output ONE most likely physical activity label for the entire sequence. Possible physical activity classes are: ``walking'', ``walking\_upstairs'', ``walking\_downstairs'', ``sitting'', ``standing'', ``laying''.

You must base your decision on the actual acceleration values shown in the plot. You are required to identify and report the approximate value ranges (minimum and maximum) of the x, y, and z channels over the time period. Do not make a classification based solely on the shape or variability of the waveform---your reasoning must include explicit reference to the value ranges and how they support your classification.
}

\end{tcolorbox} } \end{figure}

\
\subsection{Instruction Following}

To ensure interpretable and consistent outputs, we provide models with a structured instruction format applicable across all datasets. The instruction enforces step-by-step reasoning and standardized label selection in a consistent format.
\vspace{1em}
\begin{figure}[H] \centering \resizebox{1.0\textwidth}{!}{ \begin{tcolorbox}[colback=purple!5!white, colframe=purple!70!black,
title=Instruction Format for Reasoning and Classification, boxrule=0.3mm, width=\textwidth, arc=3mm, auto outer arc=true]
{\ttfamily
You MUST first reason step-by-step about the observed time-series patterns and task-specific signal characteristics. This reasoning process MUST be clearly written in natural language.

After completing your reasoning, you must select ONE most appropriate label from the predefined class list and clearly state your final prediction.

[class\_1], [class\_2], ...
}
\end{tcolorbox} } \end{figure}

If structured output is required for external evaluation, such as soft reward scoring using an LLM-as-a-Judge, we additionally enforce a clear three-part format in natural language, as shown below:
\vspace{1em}
\begin{figure}[H] \centering \resizebox{1.0\textwidth}{!}{ \begin{tcolorbox}[breakable,
  colback=purple!5!white, 
  colframe=purple!70!black,
  title={Instruction Format for Reasoning, Classification, and Extension},
  boxrule=0.3mm, 
  width=\textwidth, 
  arc=3mm, 
  auto outer arc=true
]

{\ttfamily
You MUST first reason step-by-step about the observed time-series patterns and task-specific signal characteristics.  
This reasoning process MUST be clearly written in natural language.

After completing your reasoning, you must select ONE most appropriate label from the predefined class list and clearly state your final prediction:  
[class\_1], [class\_2], ...

Finally, provide an additional recommendation, suggestion, or interpretation based on your reasoning and prediction.  
This part MUST be enclosed within \texttt{<extension>} and \texttt{</extension>} tags.
}
\end{tcolorbox} } \end{figure}

\subsection{Prompt for Soft Reward}

To assess the quality of open-ended model-generated suggestions (\texttt{<extension>} outputs), we adopt a soft reward mechanism based on the LLM-as-a-Judge paradigm. Specifically, we employ GPT-4o via the OpenAI API as a frozen expert model to evaluate each extension along four semantic dimensions. The judge returns a continuous score between  0.0 and 1.0, representing the averaged quality across the criteria below.

\begin{figure}[H] \centering \resizebox{1.0\textwidth}{!}{ \begin{tcolorbox}[breakable,colback=blue!3!white, colframe=blue!60!black,
title=General Soft Reward Prompt (LLM-as-a-Judge), boxrule=0.3mm, width=\textwidth, arc=3mm, auto outer arc=true]
{\ttfamily
You are an expert assistant evaluating a model-generated reasoning, classification, and recommendation output. The goal is to assess the quality of the recommendation (i.e., extension) based on the following structured output:

\textbf{Reasoning:} [reasoning] \\
\textbf{Prediction:} [class\_X] \\
\textbf{Extension:} [extension]

Score the extension based on the following four criteria, each on a scale from 0.0 to 1.0. Then return \textbf{only the average score} as a single float (e.g., \texttt{0.625}).

\textbf{Evaluation Criteria:} \\
1. \textbf{Specificity} – Is the extension clearly tailored to the predicted class or context, avoiding generic language? \\
2. \textbf{Appropriateness} – Is the recommendation suitable given the prediction and reasoning? \\
3. \textbf{Relevance} – Does the extension logically follow from the preceding reasoning and predicted class? \\
4. \textbf{Depth} – Does the extension demonstrate domain knowledge, including detailed actions or next-step considerations?

\textbf{Instructions:} \\
- Penalize vague or generic outputs (e.g., "be careful") unless well-justified. \\
- Reward informative, actionable, and context-aware suggestions. \\
- Return only a single float between 0.0 and 1.0.
}
\end{tcolorbox} } \end{figure}

\section{Warm-Up Data Collection via GPT-4o}\label{warm-up}

To construct high-quality time series–text paired reasoning data for warm-up training, we adopt an automated pipeline leveraging GPT-4o via the OpenAI API (with temperature set to 1.0 and all other parameters at default values). Using our structured prompt template for reasoning and classification with instruction-following, each sample is processed by GPT-4o to generate a complete reasoning–classification pair. Only samples whose predicted labels match the ground truth are retained as high-quality supervision. 
We collect approximately 1,000 class-balanced samples per dataset. For datasets with limited available instances (TEE, EMG, CTU), fewer samples are included. All collected data will be publicly released upon publication to support future research. Below, we present several  cases for reference.

\begin{figure}[H] \centering \resizebox{1.0\textwidth}{!}{ \begin{tcolorbox}[breakable,colback=white!5!white, colframe=green!40!black,
title=Generated Reasoning Sample (RCW), boxrule=0.3mm, width=\textwidth, arc=3mm, auto outer arc=true]

\ttfamily
[{"from": "human", "value": "Prompt for RCW + Instruction Format for Reasoning and Classification"}]
 \\
{"from": "gpt", "value": "<think> 
To classify the audio signal for the presence of a right whale call, we need to consider several characteristics: \\
1. **Frequency Range**: Right whale up-calls typically range from 60Hz to 250Hz. Since the sampling rate is 2kHz, any frequency components within this range should be detectable. \\
2. **Duration**: Right whale calls usually last around 1 second. Given the 2-second duration of the waveform, any call should be visible if it exists. \\
3. **Signal Characteristics**: The waveform should show a distinct modulation corresponding to the up-call frequency. This would likely appear as a consistent pattern or peak within the correct frequency range over the duration of the call. \\
4. **Noise Levels**: Low-frequency noise from other activities may obscure the call, making it less visible. \\
In the provided waveform: \\
- The waveform appears mostly consistent without distinct peaks that would stand out as a right whale up-call. \\
- The signal amplitude ranges within approximately -0.02 to 0.02. However, there's no clear modulation or pattern around 60Hz to 250Hz that is indicative of an up-call. \\
- The waveform seems to have a random appearance rather than showing a periodic pattern typically associated with the structured nature of animal vocalizations. \\
Given the absence of any noticeable patterns or frequencies that align with right whale calls, along with the information that the ground truth is NO WHALE, it is reasonable to conclude there is no right whale present in this segment. 
</think> \\
<class>NoWhale</class>"}]
\end{tcolorbox} } \end{figure}

\begin{figure}[H] \centering \resizebox{1.0\textwidth}{!}{ \begin{tcolorbox}[breakable,colback=white!5!white, colframe=green!40!black,
title=Generated Reasoning Sample (HAR), boxrule=0.3mm, width=\textwidth, arc=3mm, auto outer arc=true]

\ttfamily
[{"from": "human", "value": "Prompt for HAR +  Instruction Format for Reasoning and Classification"}]
 \\
{"from": "gpt", "value": "<think> 
The signals show the accelerometer readings over 2.56 seconds with 128 sampled data points. The x-axis shows prominent variations ranging approximately from about -0.5 to 1.0. This suggests a repetitive pattern typical in activities involving periodic motion like walking or walking downstairs. The y-axis values also fluctuate but with less magnitude, and range from about -0.3 to 0.6, indicating side motion generally seen during walking activities. The z-axis shows some variations as well, ranging approximately from -0.4 to 0.5, which might indicate body movement in the vertical plane. The combined range and periodic nature of these values, particularly the pronounced variation in the x-axis and moderate fluctuations in the z-axis, are more consistent with a stair descent activity than level walking. Furthermore, high peaks with moderate intervals suggest impact force experienced when stepping downward. Thus, the data likely corresponds to 'WALKING\_DOWNSTAIRS'.
</think> \\
<class>WALKING\_DOWNSTAIRS
</class>"}]
\end{tcolorbox} } \end{figure}


\end{document}